\documentclass[letterpaper, 10 pt, conference]{ieeeconf}  

\IEEEoverridecommandlockouts                              
\usepackage{graphics} 
\usepackage{epsfig} 
\usepackage{mathptmx} 
\usepackage{times} 
\usepackage{amsmath} 
\usepackage{amssymb}  

\usepackage{float}
\usepackage[fleqn]{mathtools}
\usepackage{algpseudocode}
\usepackage{algorithm}
\usepackage{varwidth}
\usepackage{subfigure}
\usepackage[english]{babel}	

\makeatletter
\def\BState{\State\hskip-\ALG@thistlm}
\makeatother
\title{\LARGE \bf
 Decoupled Sampling Based Planning Method for Multiple Autonomous Vehicles}

\author{Fatemeh Mohseni$^{1}$ and Mahdi Morsali$^{2}$
\thanks{Authors are with Electrical Engineering Department,
        Link\"oping University, 581-83 SE, Link\"oping, Sweden.
        {\tt\small $^{1}$fatemeh.mohseni@liu.se,$^{2}$mahdi.morsali@liu.se}}%
}

\begin{document}

\maketitle
\thispagestyle{empty}
\pagestyle{empty}

\begin{abstract}

This paper proposes a sampling based planning algorithm to control autonomous vehicles. We propose an improved Rapidly-exploring Random Tree which includes the definition of K- nearest points and propose a two-stage sampling strategy to adjust RRT in other to perform maneuver while avoiding collision. The simulation results show the success of the algorithm.

\end{abstract}

\section{INTRODUCTION}
Motion planning plays an important role in navigation of autonomous vehicles. In presence of constraints, such as collision avoidance, speed limits and rules of motion, it guarantees to find a trajectory from initial point to the goal point. 
In recent studies, different methods have been proposed and developed in this field. A vast introduction to motion and path planning problems and existing techniques and solutions can be found in \cite{LaValle}, \cite{16}, \cite{DRHCC}, \cite{A4}, \cite{Choset}. 
It depends on the nature of the problems that which methods are more appropriate and work better than other methods. 
For an ideal motion planner, there exist a few requirements, including computational complexity, optimality and completeness. However, few of them try and can solve the planning problem in its complete generality \cite{Latombe}.
Several heuristic search algorithms for path planning, have been proposed and used in the known workspace, such as $A^*$, $Grassfire$, $Dijkstra$, and $D^*$. There are other methods which are based on model predictive control (MPC), and allow to plan trajectories while taking in to account the complex vehicle dynamics \cite{Cairano}. For example, an algorithm based on MPC is proposed in \cite{Du} for real-time obstacle avoidance for ground vehicles. 
MPC also has been combined with motion primitives in \cite{Gray}, \cite{A3}, \cite{LaValle-2}  in order to plan controls for fast maneuvering of ground vehicles. 
For autonomous vehicle applications, where the vehicle has to move in an environment which is obstacle rich, the computational complexity of the motion planning algorithm is an important issue.
Since, the vehicles usually move at high speed, the path planner has to find a collision free path quickly. 
The computational time of complete and deterministic complete motion planning algorithms grows exponentially with the dimension of the configuration space. Hence, these algorithms usually are not appropriate for real time path planning problems for autonomous vehicles, especially for the problems that contain rich obstacles.
Furthermore, the optimal path of a vehicle may become infeasible due to different static and dynamic obstacles. Therefore, if during the high-speed movement of a vehicle, a preplanned trajectory becomes infeasible, multiple candidate trajectories are required. An alternative technique for these situations is to use sampling-based algorithms. 
Recently, probabilistic sampling-based methods, such as rapidly exploring random trees algorithm (RRT) \cite{Kazemi}, probabilistic roadmap algorithm (PRM) \cite{Kavraki} and PRM$^*$ \cite{DARPA}, have been proposed and developed for robot and vehicle path planning.  
These sampling based algorithms made it possible to solve motion planning problems that was considered infeasible before \cite{Choset} especially in high dimension and complex environments.
In these algorithms, instead of requiring to have an explicit expression of the configuration space, a roadmap which is a topological graph is constructed which represents the path alternatives.
Although these sampling based algorithms are not complete, they provide probabilistic completeness to ensure planning as successful as possible. 
When there is at least one feasible path, as the number of sampling nodes tends to infinity, the probability of failure of the algorithm to find a feasible path will exponentially decay to zero. However, the selection of random node leads to different planning costs. Based on this, in recent days, different asymptotically optimal RRT-based path planning algorithms were proposed in \cite{Goretkin} - \cite{Karaman}, and \cite{Lee}, \cite{Jeon}.
It has been shown that for RRT and other sampling-based path planners, the workspace is explored efficiently only when this planning cost function reflects the true cost-to-go \cite{Cheng}.
As it has been shown in \cite{LaValle-2}, the choice of a distance metric as cost function to find the nearest node affects the performance of RRT-based algorithms significantly.
For our motion planning problem, we proposed two RRT based algorithms, one for path planning and the other one for motion timing in order to avoid collision between different vehicles.  
In the past decade, the rapid development in the field of autonomous vehicles, going from single vehicle tasks to missions that require cooperation, coordination, and communication among a number of vehicles, makes the availability of adaptable motion planners more and more important.
When a group of vehicles are tasked to carry out a mission in a cooperative way in presence of complex obstacles, the inter vehicle collision adds to the complexity of the mission planning systems.
In such a systems, it is also required to guarantee that each vehicle meets spatial configuration constraints.
Several approaches have been suggested to solve the motion planning problem of multiple autonomous agents. 
More specifically, in \cite{Baskar} a vehicle-follower control based on model-based predictive control is proposed; in \cite{Ferrara} a sliding mode longitudinal controller is used to control a group of vehicles which have inter-vehicle communication; \cite{Guvenc} suggested a cruise control, in which vehicle uses information about the spacing and the relative speed from the following and the preceding vehicles; in \cite{Oncu}, a system of the automatic vehicle following has been suggested to adopt a constant spacing policy; and \cite{Rudwan} suggested a cruise control algorithm using fuzzy concept. A game theory based approach is described by \cite{Swaroop}, \cite{10}, \cite{11} and \cite{Ploeg} to guarantee safety during the maneuver for all vehicles. \cite{Yang} suggested a static game approach for lane merging maneuver.\\
When more than one autonomous vehicle work in the same area, the problem of vehicle collision has to be faced. Even if the mission space is planned and cleared of any conflict between cars, it may happen that vehicles collide. The collision can be due to different dynamic and kinematic characteristics, speed and external disturbances. Therefore, a key issue for a multivehicle maneuver is safety, represented by the requirement that cars never collide. Consequently, an approach that controls multiple autonomous vehicles, with a collision avoidance feature, becomes a way to improve the transportation system.
In this paper, we address the multiple-vehicle motion planning problem by dividing it into two phases: 1) planning path for each vehicles by proposing and using an improved RRT method which increases the optimality property of the standard RRT and decreases the computational time and 2) motion timing phase by proposing a semi deterministic sampling method which guarantees the collision avoidance between different vehicles. \\
The remainder of the paper is organized as follows. In Section \ref{Sec.Formulation}, the problem formulation is defined including our decoupled planning method definition, the vehicle model, and integration method analysis. Then, the IRRT algorithm for path planning and the motion timing algorithm are provided in Section \ref{Sec.DSBP} followed by decoupled sampling based algorithm for multiple vehicles . In section \ref{Sec.Simulation}, simulation results are provided to show the effectiveness of the proposed approach. Concluding remarks are given in Section \ref{Sec.Conclusion}.\\
\begin{table}[hb]
  \begin{center}
  \small
  \caption{??.}\label{Tab.ODE}%
  \begin{tabular}{|c|c|c|}
  \hline
 ??& ??& ??\\\hline
  ? & ?& ?\\\hline
  ? & ?& ?\\\hline
  ? & ?& ?\\\hline
  ? & ?& ?\\\hline
  ? & ?& ?\\\hline
  ?& ?& ?\\

\hline
\end{tabular}
\end{center}
\end{table}
\section{Problem formulation}\label{Sec.Formulation}
\subsection{Vehicle Dynamics}
\begin{figure}
	\begin{center}
		\includegraphics[width=8cm]{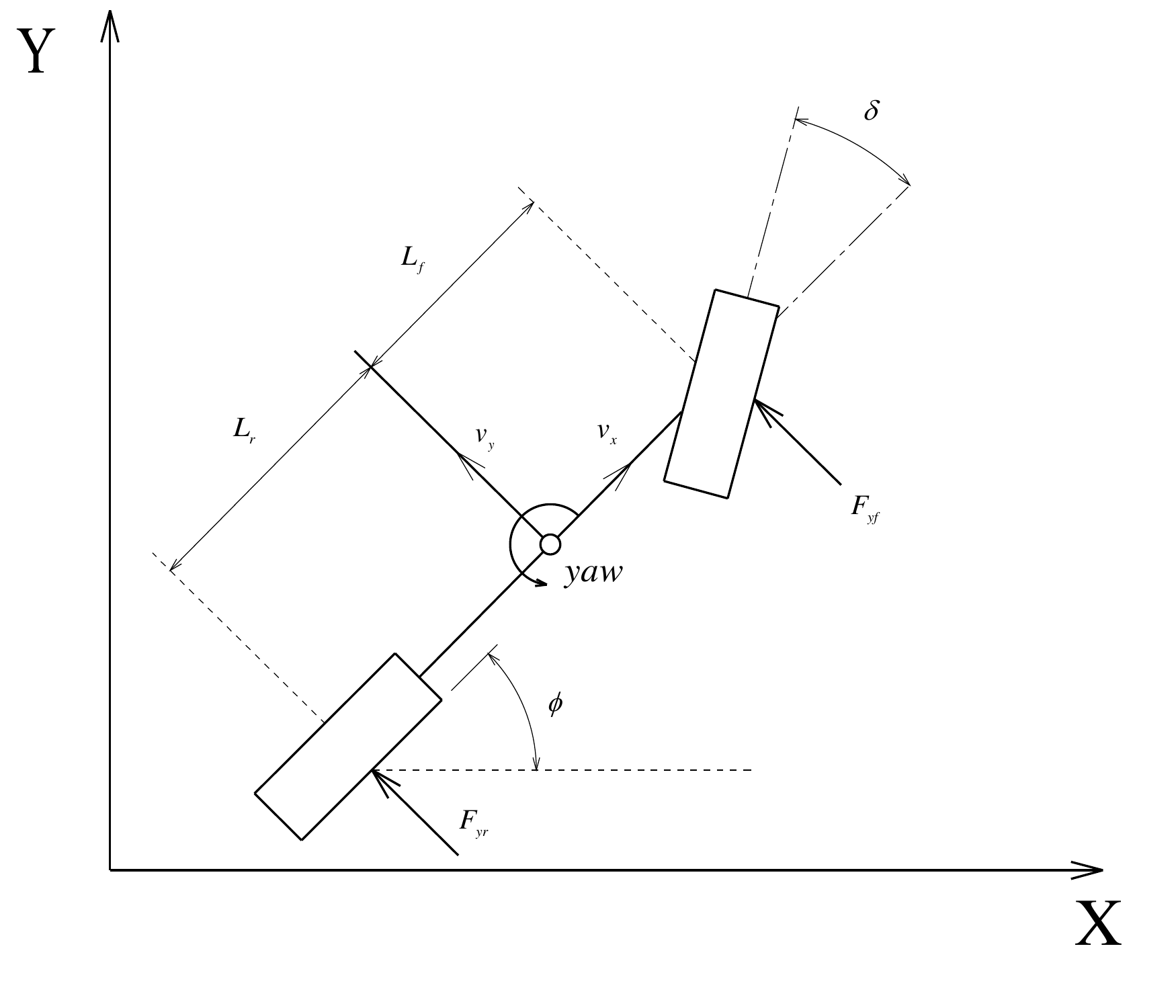}    
		\caption{Parametric notations related to vehicle kinematics, bicycle model.}
		\label{Fig.Vehicle dynamic}
	\end{center}
\end{figure}

It is assumed that movement of each vehicle is described by bicycle model which is an ordinary differential equation (ODE) given by
\begin{align}
&{\dot X }= {v_x.cos(\theta) - v_y.sin(\theta)}\notag \\
&{\dot Y} = {v_x.sin(\theta) + v_y.cos(\theta)}\notag\\
&{\dot\theta} = {r}\notag\\
&{\dot v_x }= {\frac{-F_{xf}}{m}.cos(\delta) -  \frac{F_{yf}}{m}.sin(\delta) - \frac{F_{xr} }{m}+ v_y.r}\\
&{\dot v_y }= {\frac{F_{yf}}{m}.cos(\delta) -  \frac{F_{xf}}{m}.sin(\delta) + \frac{F_{yr} }{m} - v_x.r}\notag\\
&{\dot r }= {\frac{L_f}{I_z}(F_{yf} cos(\delta) - F_{xf} sin(\delta)) -\frac{L_r}{I_z}F_{yr}}\notag
\end{align}
where $p=(X,Y)$ is the Cartesian coordinates of the vehicle's center, $\theta$ is the orientation angle, $v_x$ and $v_y$ are longitudinal and lateral speeds, respectively, $r$ is the yaw rate, $\delta$ is steer angle, $L_f$, $L_r$ are ****** and $I_z$ is******,  see Fig. \ref{Fig.Vehicle dynamic}. In addition , $F_{xf}$, $F_{xr}$ are the longitudinal and $F_{yf}$, $F_{yr}$ are the lateral forces acting on front and rear wheel that are given by
\begin{equation}\label {Forces}
\begin{cases}
	{F_{yf}} = {-C_{\alpha f}.\alpha_f}\\
	{F_{yr}} = {-C_{\alpha r}.\alpha_r}
\end{cases}
\end{equation}
where $\alpha_f$, $\alpha_r$ are slip angles of front and rear wheels respectively and described by
\begin{equation}\label {alpha}
\begin{cases}
{\alpha_f} = {\frac{v_y+L_fr}{v_x}-\delta}\\
{\alpha_r }= {\frac{v_y - L_fr}{v_y}}
\end{cases}
\end{equation}

For simplicity, it is assumed that longitudinal speed is constant and there is no aerodynamic forces. Therefore, the vehicle dynamics is reformulated as 

\begin{equation}\label {Final Vehicle Dynamic}
\begin{cases}
{\dot X }= {v_x.cos(\theta) - v_y.sin(\theta)} \\
{\dot Y }={ v_x.sin(\theta) + v_y.cos(\theta)}\\
{\dot \theta }={ r} \\
{\dot v_y }= {\frac{F_{yf}}{m}.cos(\delta) -  \frac{F_{xf}}{m}.sin(\delta) +\frac {F_{yr}}{m} - v_x.r}\\
{\dot r }= {\frac{L_f}{I_z}(F_{yf} cos(\delta)  - \frac{L_r}{I_z}F_{yr}}
\end{cases}
\end{equation}
\subsection{Integration method}
\begin{figure}[H]
   
      \includegraphics[scale=0.9]{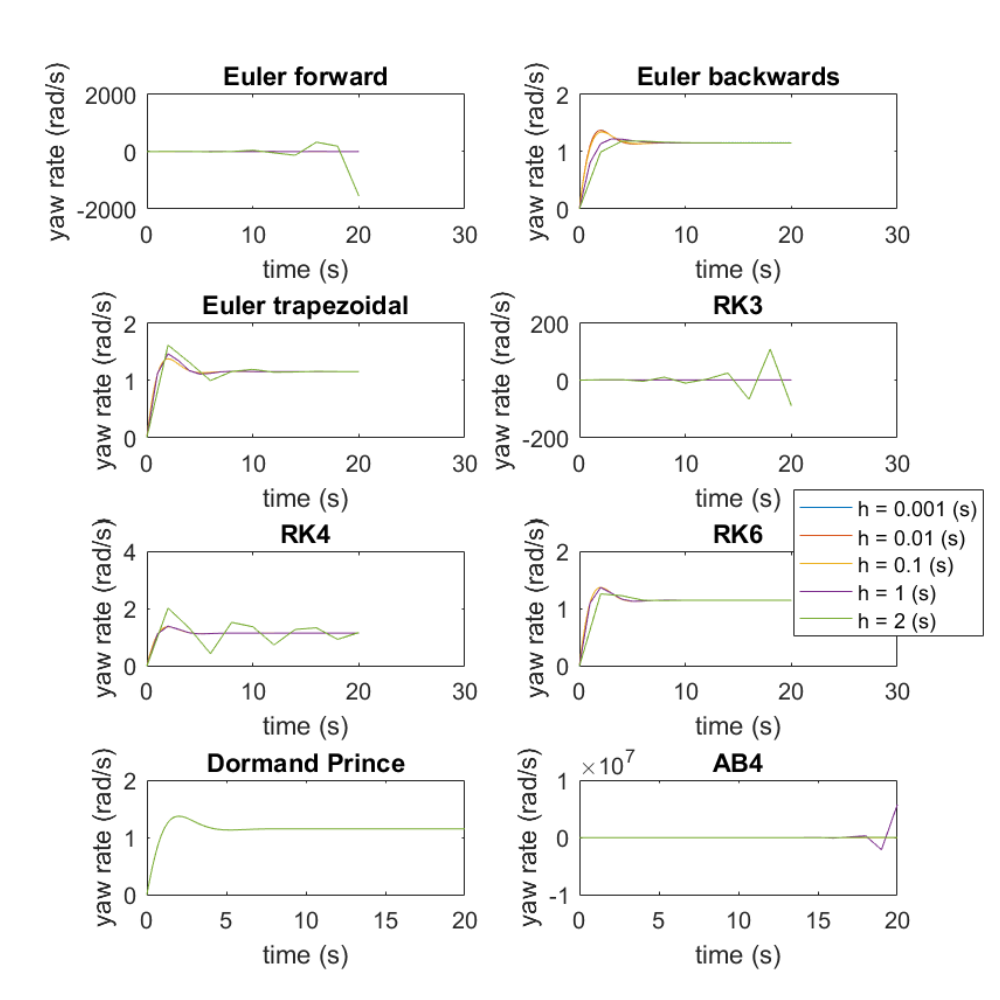}
      \caption{Stability test of integration methods in calculation of yaw rate.}
      \label{RK_YAW}
   \end{figure}

The vehicle dynamics represented by bicycle model, should be integrated at each iteration during the search algorithm. In order to have reliable results, the integrator should be accurate, stable and fast enough. For this purpose, the bicycle dynamic described by (\ref{Final Vehicle Dynamic}) is integrated by using different methods in finite time. The integration methods that are used during the test are, Euler Forward, Euler Backwards, Trapezoidal, 3rd, 4th, 6th, order Runge-Kutta, Dormand-Prince and 4th order Adams-Bashforth methods. \\
The test was performed with a constant longitudinal speed of $15 m/s$ and a steer angle of $pi/4$.The Dormand-Prince method is an adaptive method that is used to illustrate real values of the parameters and make comparison with other methods.
  \begin{table}[hb]
  \begin{center}
  \small
  \caption{Error and computational time for different integration methods.}\label{Tab.ODE}%
  \begin{tabular}{|c|c|c|}
  \hline
  Method& Error& computational time\\\hline
  ? & ?& ?\\\hline
  ? & ?& ?\\\hline
  ? & ?& ?\\\hline
  ? & ?& ?\\\hline
  ? & ?& ?\\\hline
  ?& ?& ?\\
    ?& ?& ?\\
\hline
\end{tabular}
\end{center}
\end{table} 
The results showed that Euler Forward, 3rd order Runge-Kutta and Adams-Bashforth methods are not stable for large step sizes. Among the other methods, the 4th order Runge Kutta shows a good stability for even large step sizes and as a trade off between accuracy, computational effort and stability the 4th order Runge Kutta is a good candidate for this problem, see \ref{RK_YAW}.
Table \ref{Tab.ODE} illustrates the error and computational time for each method. 
  \begin{table}[hb]
  \begin{center}
  \small
  \caption{Error and computational time for different integration methods.}\label{Tab.ODE}%
  \begin{tabular}{|c|c|c|}
  \hline
  Method& Error& computational time\\\hline
  ? & ?& ?\\\hline
  ? & ?& ?\\\hline
  ? & ?& ?\\\hline
  ? & ?& ?\\\hline
  ? & ?& ?\\\hline
  ?& ?& ?\\
    ? & ?& ?\\\hline
    ? & ?& ?\\\hline
    ? & ?& ?\\\hline
    ? & ?& ?\\\hline
    ? & ?& ?\\\hline
    ?& ?& ?\\
\hline
\end{tabular}
\end{center}
\end{table}
  
\section{Decoupled sample based method for multiple vehicles motion planning}\label{Sec.DSBP} 
\subsection{Improved RRT algorithm for path planning}
In IRRT algorithm, the definition of parents in standard RRT is modified in order to get smoother path and also use most likely each random point. To do that, after selecting each random point, the nearest point to that in the tree is found. Then the K-nearest vector is defined as follows\\

\emph{Definition 1, (K-nearest vector)}:  K-nearest vector,  $P_{near}=(p_{near1},\dots,p_{neark})^T$ for each random point is defined as its nearest point in the tree, parent of that nearest point and $K-2$ parents of that parent.\\
After finding  $P_{near}$, the  proper  steer  angles  to  move from each component of that in  the direction of sample point is calculated and fed into the vehicle dynamics.
By using the steer  angles,  the integration  takes  place  for  a  given  amount  of  time  horizon which results in $P_{new}=(p_{new1},\dots,p_{newk})^T$. For all the integrated points, $P_{new}$, a cost is calculated as follow\\

\emph{Definition 2, (cost function)}:  for each $i^{th}$ component of $P_{new}=(p_{new1},\dots,p_{newk})^T$, the cost function is defined as follow
\begin{equation}\label{cost}
F(i)=g(p_{init},P_{new}(i))+H(P_{new}(i),p_{goal})
\end{equation}

where $g$ is the traveled distance from start point to $p_{newi}$ and heuristic function $H$ is the direct distance from $p_{newi}$ to destination point. \\
For each  integrated  point, $p_{newi}$, if there is a collision between the path from $p_{neari}$ to $p_{newi}$ and obstacles, the heuristic cost, $H(i)$, is set to infinity.
The next step is to select the best $p_{newi}$. $p^*_{new}$ is the component of $P_{new}$ which does not collide obstacles and has minimum cost value. \\
While the road map has not been constructed, the $p^*_{new}$ has not arrived the terminal zone, a random value, $\rho$, will be generated. If this random random value is larger than the specified probability coefficient, $\rho'$, then $p_{rand}$ will be generated randomly inside the terminal zone; otherwise, it will be generated randomly in whole space. This biasing method has been used to increase the convergence of the method toward the destination point. After this step, the nearest node,  $p_{near}$, from $x_{rand}$ is found. By using $p_{near}$ and $p_{rand}$, the K-nearest vector, $P_{near}$ is found according to Definition 1. $f(p(i),u(i))$ is then calculated according to $P_{near}(i)$ and $p_{rand}$. 
IRRT algorithm is shown in Table \ref{IRRT}.  
\begin{center}
\begin{algorithm}
\caption{IRRT Algorithm}\label{IRRT}
    \begin{algorithmic}[1]
\Function{Path Finder}{}
\State $\text{Tree-init}(p_{start},p_{goal},~\rho' \in [0,1],~K=4,~a=10) $
\While {$!$flag do}{
        \State Generate a $\rho \in [0,1]$
        \If {$\rho \geq \rho'$}
            \State $p_{rand}=(p_{goal}(1)+a+\rho . (2.a),p_{goal}(2)-a+\rho . (2.a))^T$
        \Else
             \State $p_{rand}\leftarrow \text{SamplePoint}()$
        \EndIf
        \State $p_{near}=\text{Nearest}(G=(V,E),p_{rand})$
        \State $P_{near}\leftarrow\text{K-near}(p_{near},K)$
        \State $P_{new}\leftarrow\text{Steer}(P_{near},p_{rand})$
        \For{$i = 1$ to ${K}$}
        \If {$!\text{Collision-free}(P_{new}(i),P_{near}(i)$}
                    \State $P_{new}(i) \text{ is eliminated from } P_{new}$
                \Else
                     \State $P_{new}(i) \text{ remains in } P_{new}$
                \EndIf
                        \EndFor
                        \For{$j = 1$ to ${Numel(P_{new})}$}
                                    \State $F(P_{near}(i))=g(P_{near}(i))+h(P_{near}(i)) \text{ according to Deffinition 2}$
                                \EndFor
            \State $n=\text{arg min}_\{i\}F(i)$
            \State $p^*_{near}=P_{near}(n)$
            \State $p^*_{new}=P_{new}(n)$
        \State $V\leftarrow V\cup \{p^*_{new}\}$
                    \If {$\|p^*_{new}-p_{goal}\|\leq a$}
                                \State $\text{return flag $=$ true}$
            \Else
                                \State $\text{return flag $=$ false}$
            \EndIf
        }
        \EndWhile
\EndFunction
\end{algorithmic}
\end{algorithm}
\end{center}
\begin{figure}
\centering
\subfigure[]{\includegraphics[width=.49\columnwidth]{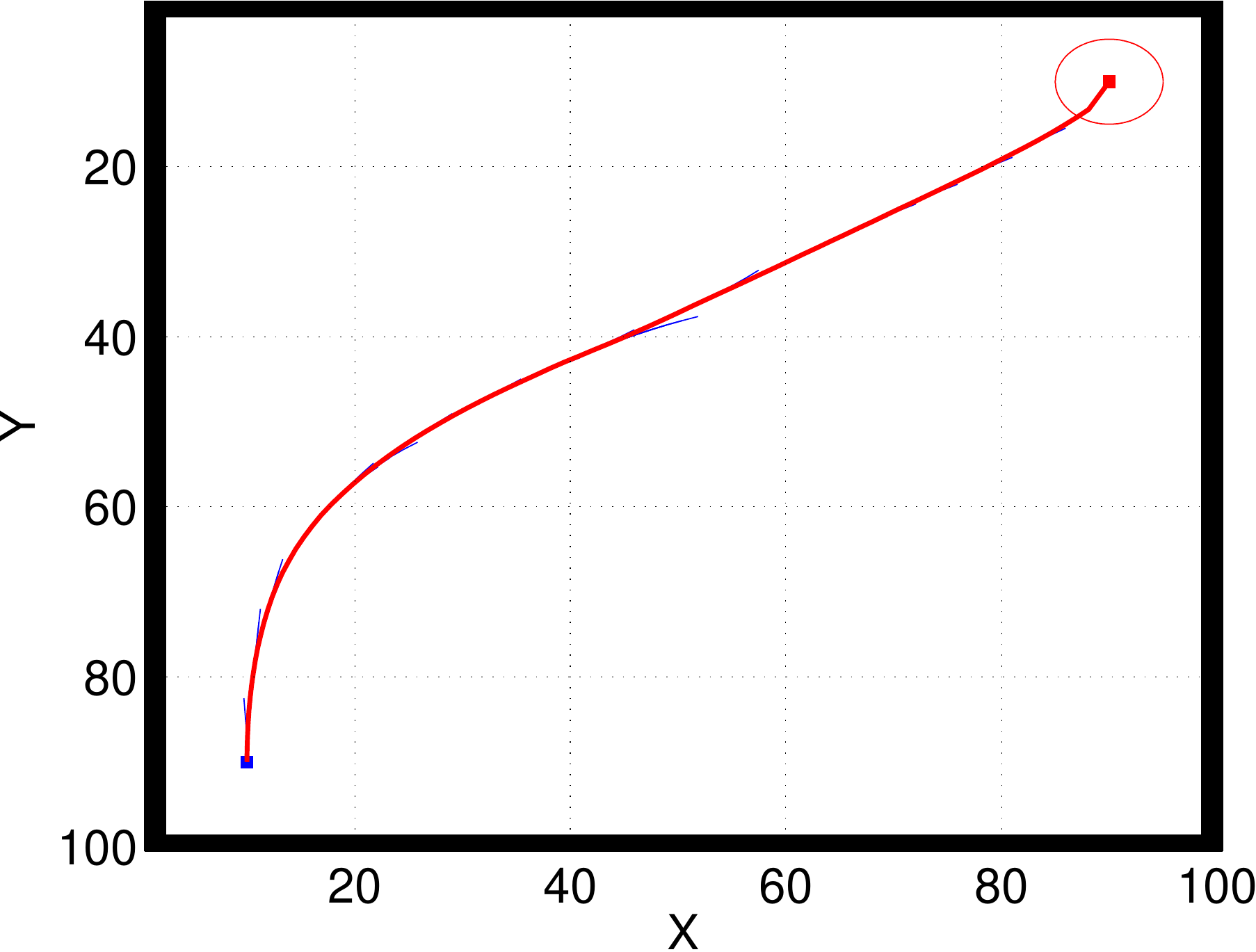}}
\subfigure[]{\includegraphics[width=.49\columnwidth]{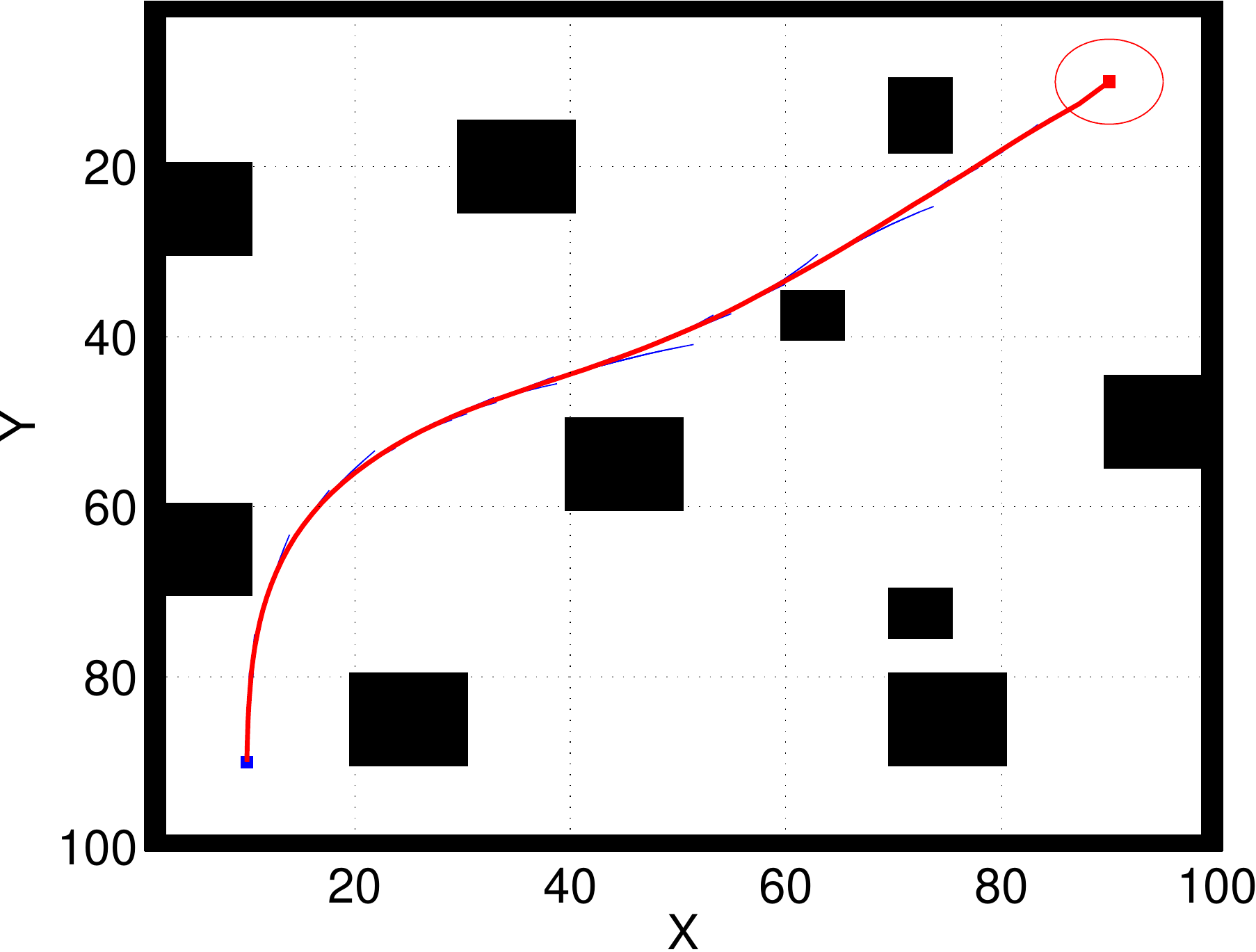}}
\subfigure[]{\includegraphics[width=.49\columnwidth]{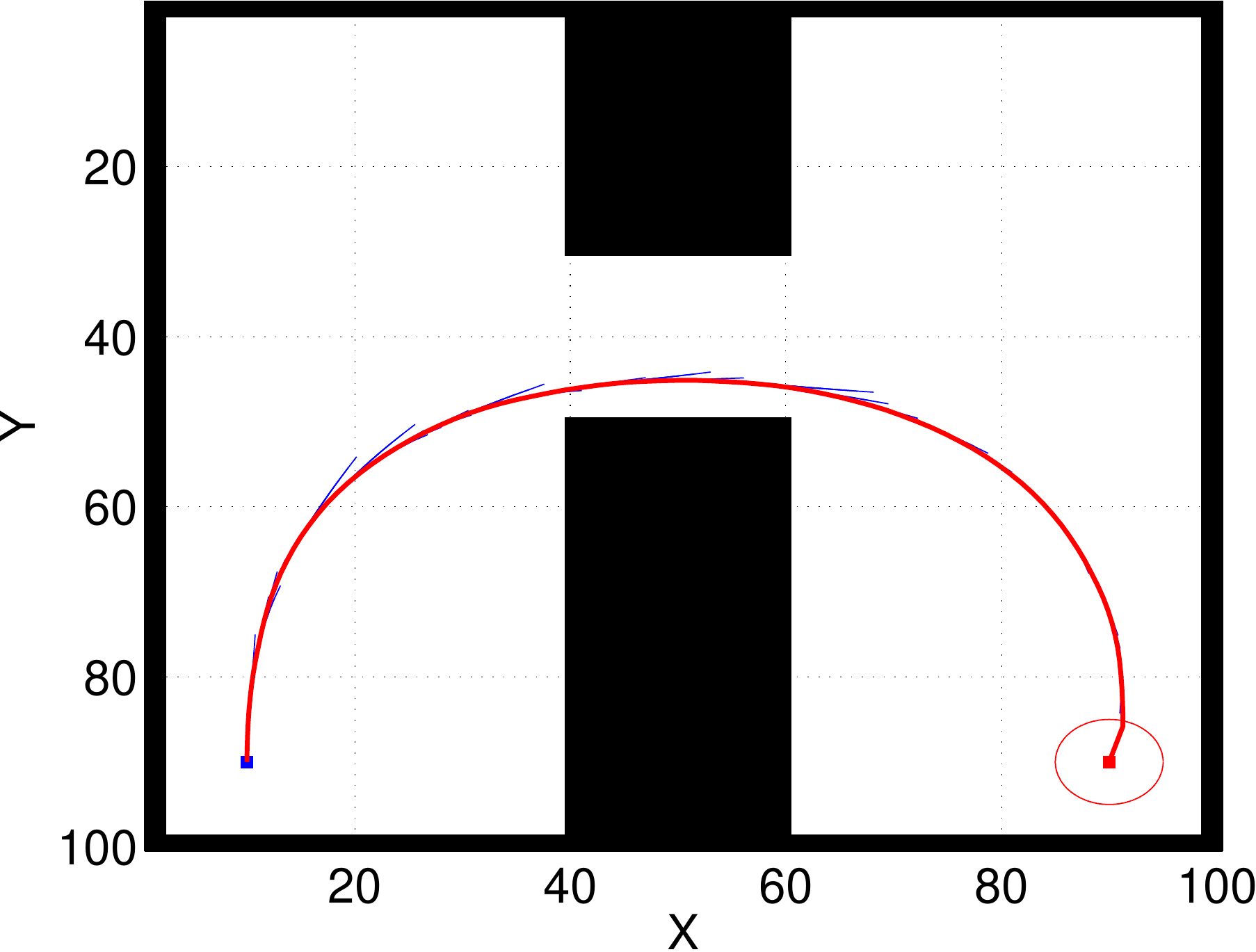}}
\subfigure[]{\includegraphics[width=.49\columnwidth]{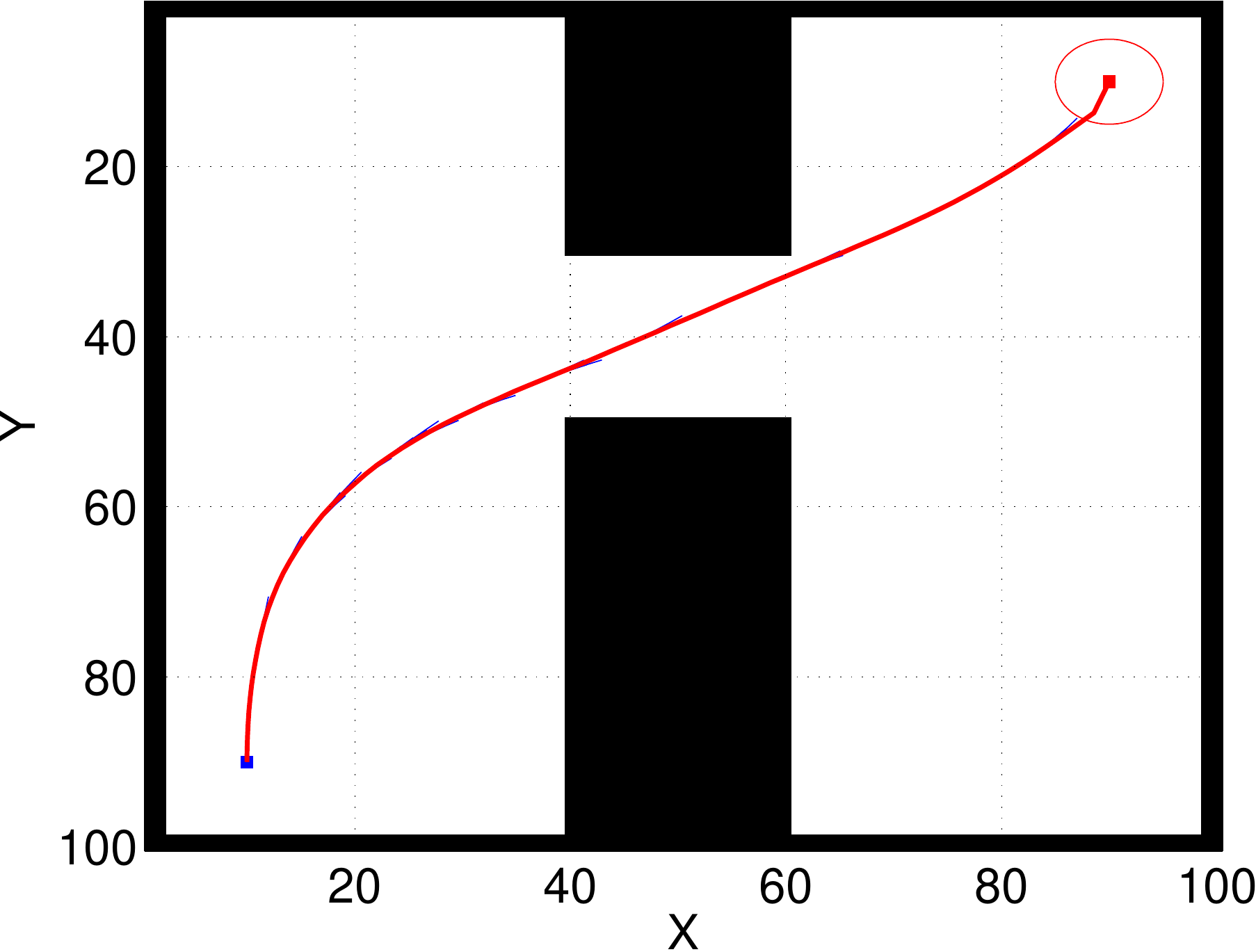}}
\subfigure[]{\includegraphics[width=.49\columnwidth]{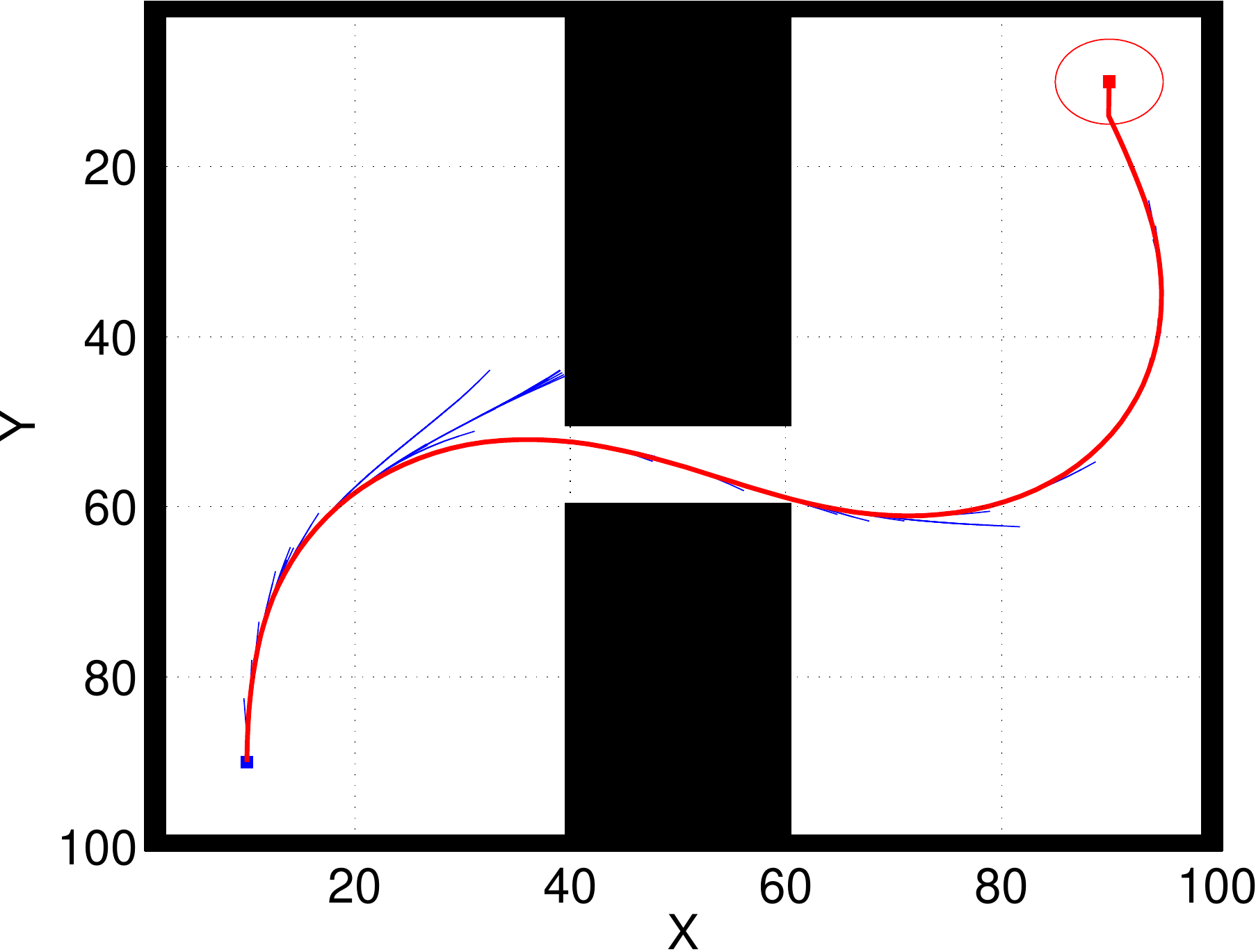}}
\caption{Vehicle path is generated using Improved RRT algorithm in different road maps.
 }
\label{Fig.IRRT}
\end{figure}

 IRRT algorithm is used for different road maps and different start and goal points, see Fig \ref{Fig.IRRT}. The results show that, the path from IRRT algorithm is close to optimal path and also the computational time is much less than RRT. Table ??? shows the computational time for different cases. 
 
  In Fig. \ref{Fig.Time}, the average of computation time of both algorithms is illustrated. Obviously, the IRRT algorithm is more efficient specially for bigger numbers of vehicles. \\
  \begin{table}[hb]
  \begin{center}
  \small
  \caption{Computational time}\label{Tab.Time}%
  \begin{tabular}{|c|c|c|}
  \hline
  & IRRT& RRT\\\hline
  ? & ?& ?\\\hline
  ? & ?& ?\\\hline
  ? & ?& ?\\\hline
  ? & ?& ?\\\hline
  ? & ?& ?\\\hline
  ?& ?& ?\\
    ? & ?& ?\\\hline
    ? & ?& ?\\\hline
    ? & ?& ?\\\hline
    ?& ?& ?\\
\hline
\end{tabular}
\end{center}
\end{table}

\subsection{Motion timing sample based method}
In the motion timing part,  a timing function, $\sigma : T \rightarrow[0,1]$, is designed for each vehicle, based on its priority. It has been assumed that some collision free path $\tau : [0, 1] \rightarrow C_{free}$ has been computed already using Algorithm 1. $\sigma$ indicates the location of the vehicle along the path, $\tau$,  at time $t$. By defining the composition $\phi = \tau \circ \sigma$, which
maps from $T$ to $C_{free}$ via [0, 1], $\sigma$ is achieved. Hence, $\phi (t)=\tau(\sigma(t))$ indicates the configuration at time $t\in T$. We assume that the vehicle with the highest priority moves along its path which was found in part 1 with constant speed. For the other vehicles, after finding the path using Algorithm 1, the configuration $\phi (t)$ should be calculated. Each vehicle should move along its path from $\tau(0)$ to $\tau(1)$ while an obstacle, $\mathcal{O}(t)$, moves along its path over the time interval $T$. For each vehicle, other vehicles with higher priority are considered as moving obstacles. 
Let domain of $\tau$ be denoted by $S=[0,1]$. $X=T\times S$ defines the state space in which the time $t \in T$ and the position along the path, $s \in [0, 1]$ is indicated by a point $(t, s)$. The obstacle region in $X$ is defined as 
\begin{equation}
X_{obs} = \{(t, s) \in X | A(\tau (s)) \cap O(t) = \O \}
\end{equation}
Therefore, $X_{free}$ is defined as $X_{free} =  X \setminus X_{obs}$ . The task is then  to find a path $g : [0, 1] \rightarrow X_{free}$. For this purpose, the VT algorithm is proposed. VT algorithm is a sample based method which is inspired by the RRT and A$^*$ algorithms, see \cite{LaValle}.\\

In this method instead of randomly selecting one point in the $X$, several points are selected at each step. The start and goal points are $(0,0)$, $(1,1)$ respectively. $(t,1)$ which is $s=1$ line is when the vehicle followed the whole path. In order to increase the convergence of the method, some random points will be chosen at each iteration on the line $s=1$. Other random points are selected in  ????????????.
After selecting K random points, $X_{new}=(x_{new1},\dots,x_{newk})$ will be calculated according to dynamic. In order to apply speed limitation, if the slope of the line that connects each $x_{rand}$ and $x_{newi}$ is less than $v_{min}$, the speed will consider to be $v_{min}$. If the slop is larger than $v_{max}$, then the speed will consider to be $v_{max}$.
Otherwise, the speed will be the slope of the line. \\ 
After finding $X_{new}$ , a cost function will be assigned to each of the elements of $X_{new}$. the cost function is defined as follow
\begin{equation}
F_2(i)=g_2(x_{init},X_{new}(i))+H_2(X_{new}(i),x_{goal})
\end{equation}
where $g_2(x_{init},X_{new}(i))$ is the length of the path between $x_{init}$ and each $X_{new}(i)$ and $H_2$ is the distance between $X_{new}(i)$ and $x_{goal}$. 
The next step is to select the best $x_{newi}$. $x^*_{new}$ the component of $X_{new}$ which is in $X_{free}$ and has minimum cost value.\\
This steps will repeated until $x^*_{new}$ reaches the line $s=1$.\\
?????????????????acceleration limit??????????????????other limits??????????????????????????????????
VT algorithm is shown in Table \ref{VT}.
\begin{center}
\begin{algorithm}
\caption{VT Algorithm}\label{VT}
    \begin{algorithmic}[1]
\Function{Velocity tuning}{}
\State $\text{Tree-init}(X_{init},X_{goal},~K=10) $
\While {$!$flag do}{
        \State Generate K$/2$ sample points on ????????????? and K$/2$ on $s=1$ line.
        \For{$i = 1$ to ${K}$}
        \State $X_{near}(i)=\text{Nearest}(G=(V,E),X_{rand}(i))$
        \State $X_{new}(i)\leftarrow\text{Acc}(X_{near}(i),X_{rand}(i))$
        \If {$!\text{Collision-free}(X_{new}(i),X_{near}(i)$}
                    \State $X_{new}\leftarrow X_{new} \setminus X_{new}(i)$
                \Else
                     \State $X_{new}\leftarrow X_{new}$
        \EndIf
        \EndFor
                        \For{$j = 1$ to ${Numel(X_{new})}$}
                             \State $F_2(X_{near}(i))=g_2(X_{near}(i))+H_2(X_{near}(i))$
                                \EndFor
            \State $n=\text{arg min}_\{i\}F_2(i)$
            \State $x^*_{near}=X_{near}(n)$
            \State $x^*_{new}=X_{new}(n)$
        \State $V\leftarrow V\cup \{x^*_{new}\}$
                    \If {$x^*_{new}=x_{goal}$}
                                \State $\text{return flag $=$ true}$
            \Else
                                \State $\text{return flag $=$ false}$
            \EndIf
        }
        \EndWhile
\EndFunction
\end{algorithmic}
\end{algorithm}
\end{center}
 
\subsection{Decoupled sample based method}
By using Algorithm \ref{IRRT} and Algorithm \ref{VT}, the method is defined in Table \ref{DSBP}. 
\begin{table}[hb]
  \begin{center}
  \small
  \caption{??.}\label{Tab.ODE}%
  \begin{tabular}{|c|c|c|}
  \hline
 ??& ??& ??\\\hline
  ? & ?& ?\\\hline
  ? & ?& ?\\\hline
  ? & ?& ?\\\hline
  ? & ?& ?\\\hline
  ? & ?& ?\\\hline
  ?& ?& ?\\

\hline
\end{tabular}
\end{center}
\end{table}
\begin{center}
\begin{algorithm}
\caption{DSBP Algorithm}\label{DSBP}
    \begin{algorithmic}[1]
\Function{Motion planning for multiple vehicles}{}
\State $\text{Initial position, goal position, state limits, n=number of vehicles}$
        \For{$i = 1$ to ${n}$}
                            \State Reorder vehicles number based on their priority
        \State $\tau(i) \leftarrow IRRT$ algorithm
        \EndFor
         \For{$j = 1$ to $n$}
                \State Find configuration $\phi_i(t)$
            \State $V_i\leftarrow$ VT Algorithm $(X_{1},\dots,X_{n-1})$
            \EndFor
           \EndFunction
           \end{algorithmic}
           \end{algorithm}
           \end{center}
           
\begin{figure}
\centering
\subfigure[]{\includegraphics[width=.49\columnwidth]{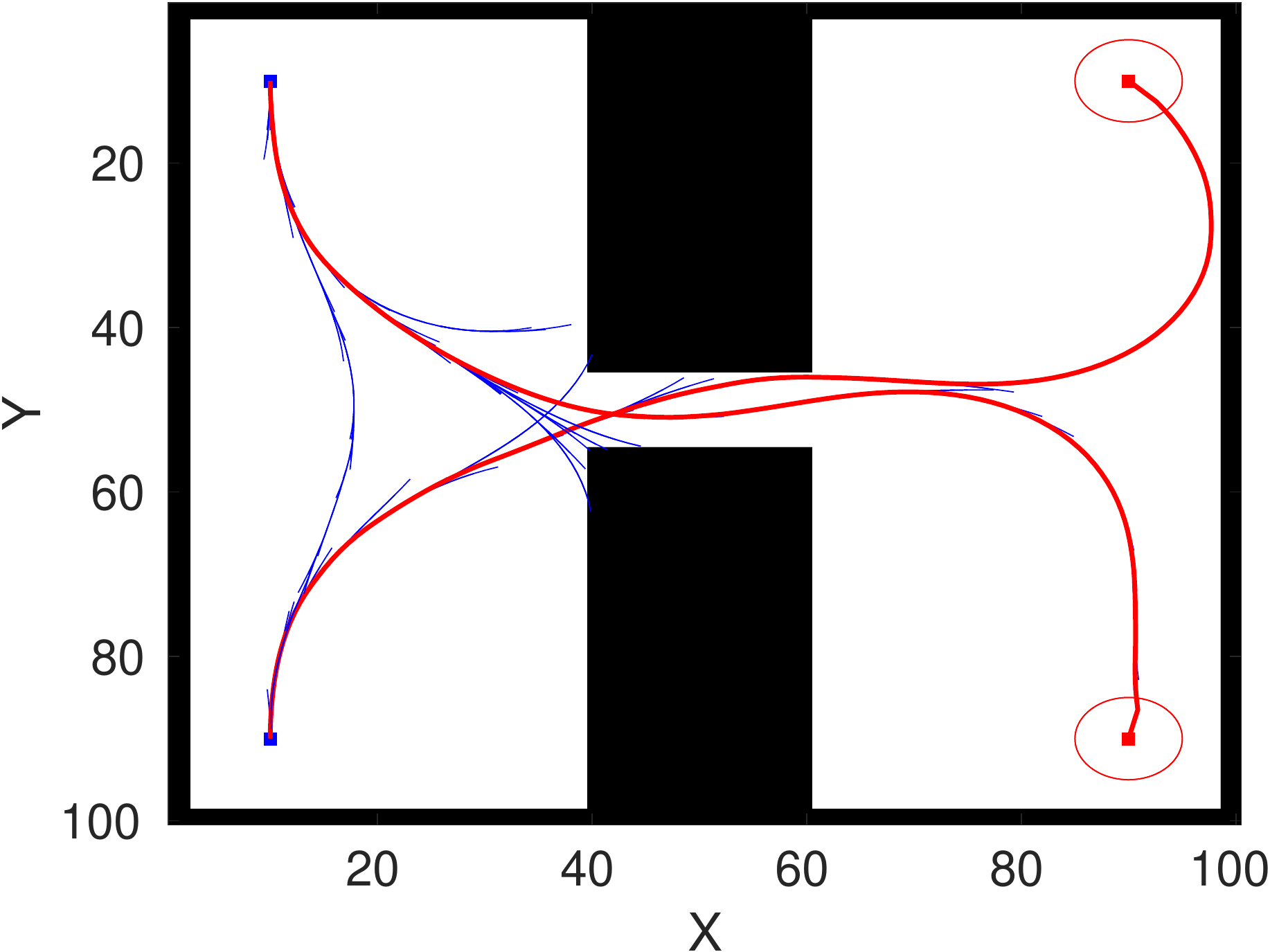}} \subfigure[]{\includegraphics[width=.49\columnwidth]{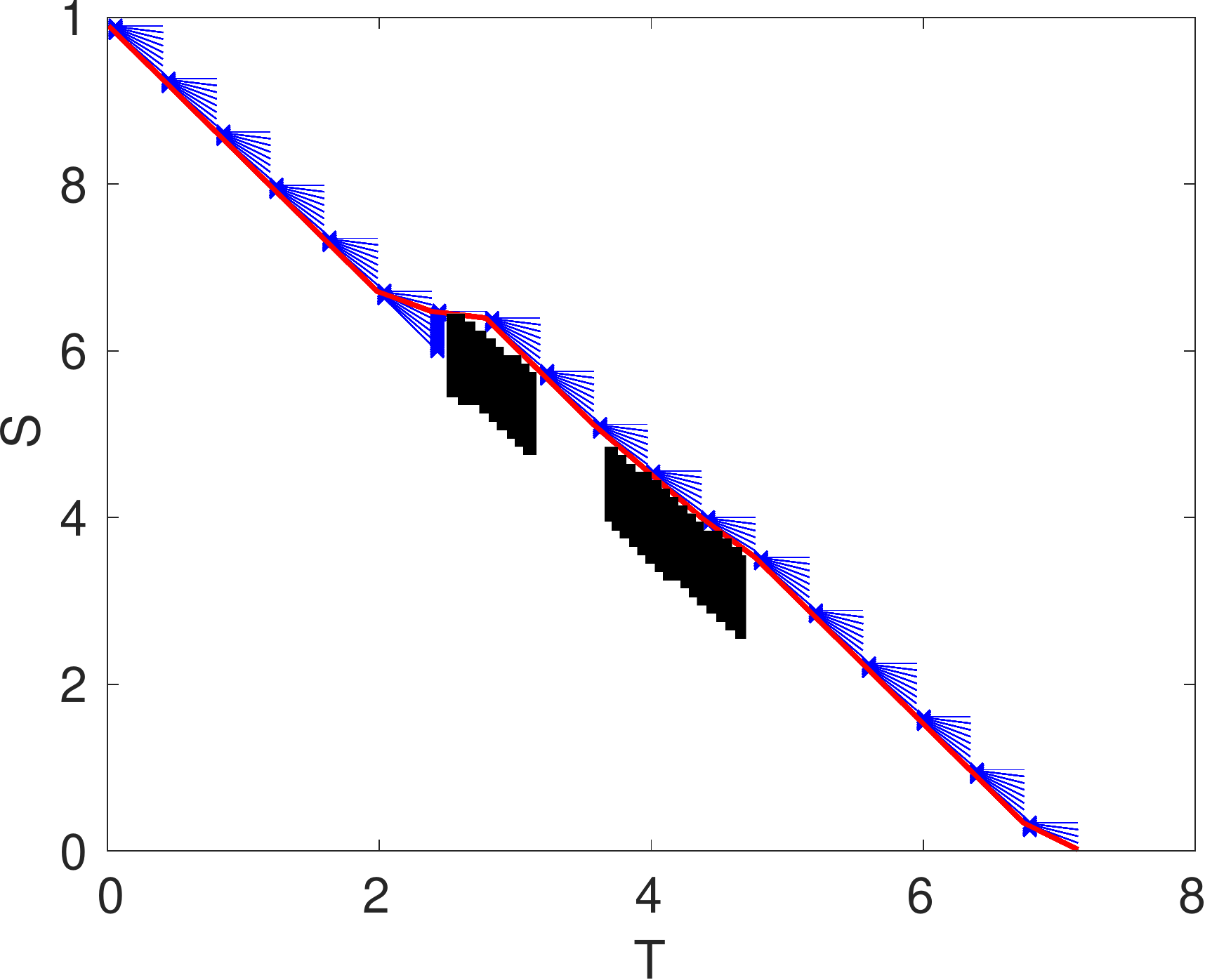}}
\subfigure[]{\includegraphics[width=.49\columnwidth]{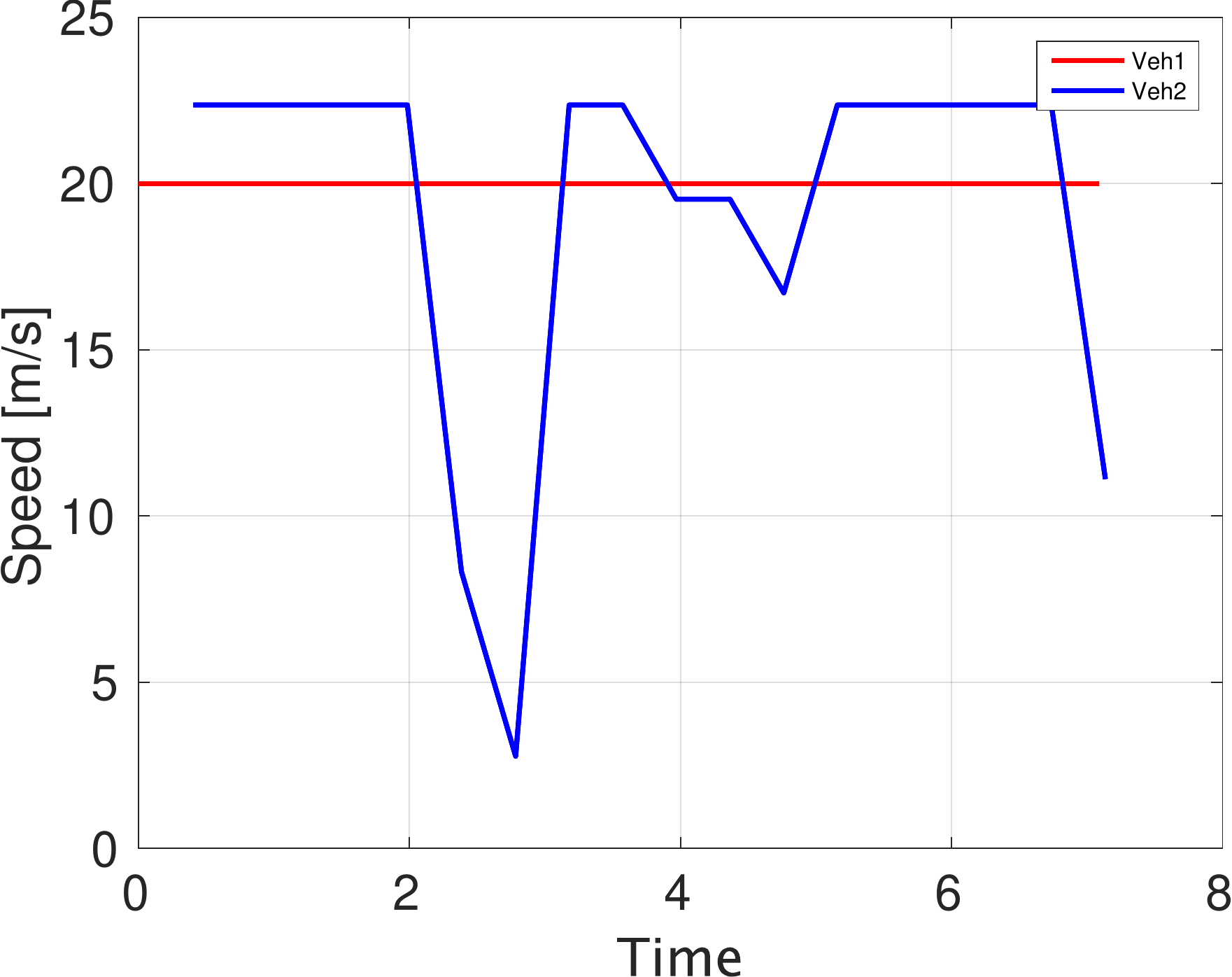}} \subfigure[]{\includegraphics[width=.49\columnwidth]{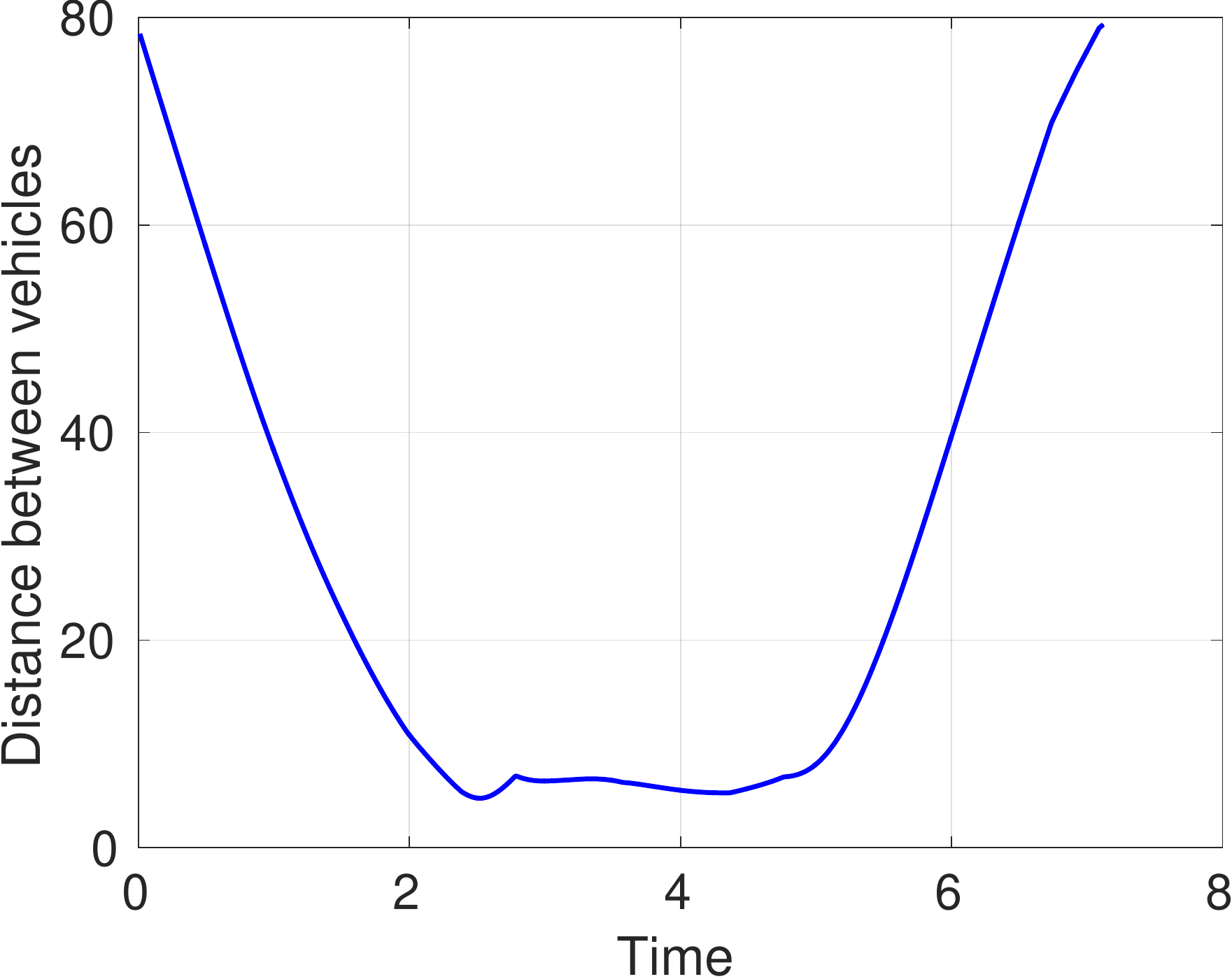}}
\caption{Simulation results of the method}
\label{Fig.Two}
\end{figure}

\begin{figure}
\centering
\subfigure[]{\includegraphics[width=.49\columnwidth]{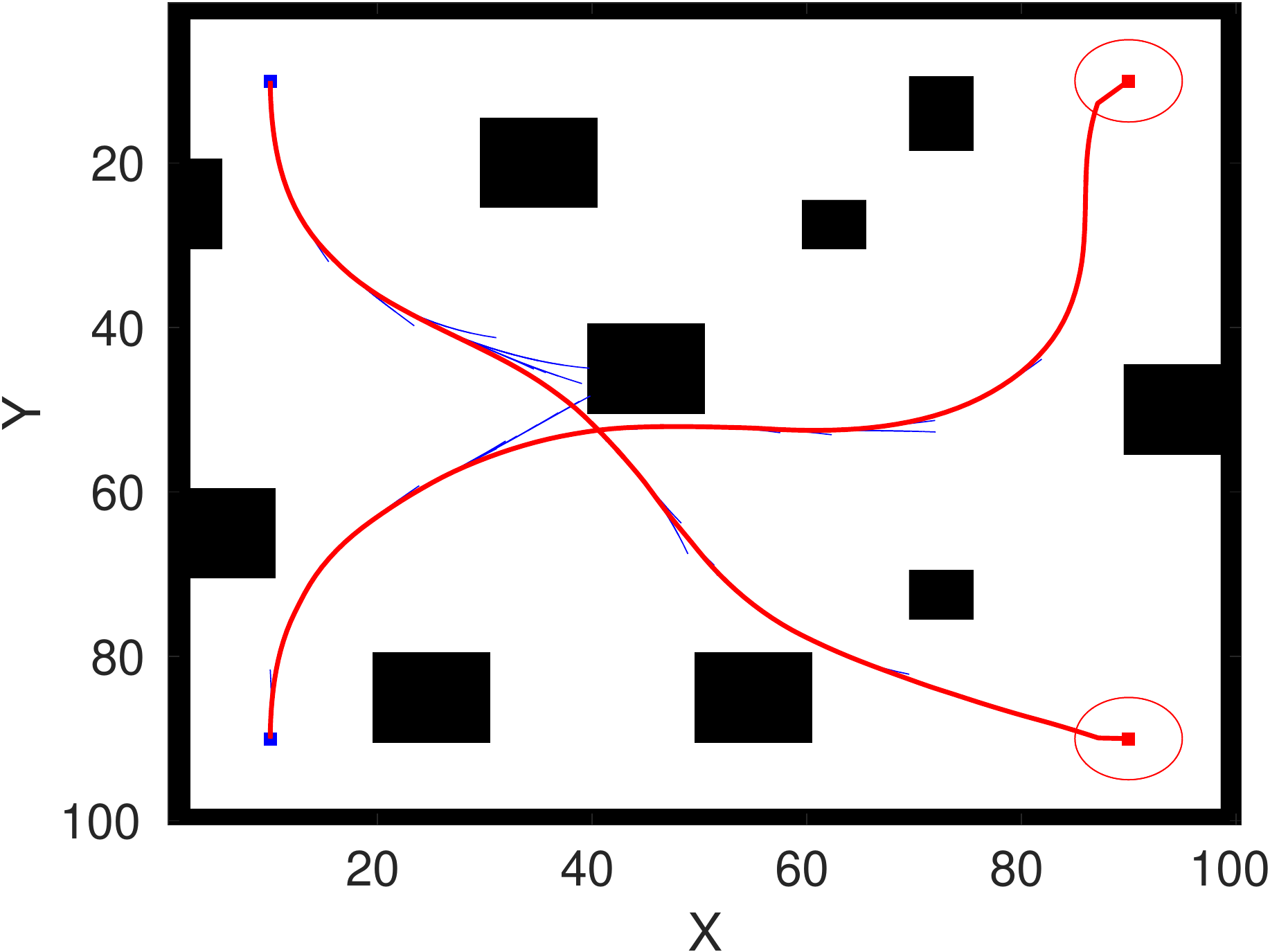}} \subfigure[]{\includegraphics[width=.49\columnwidth]{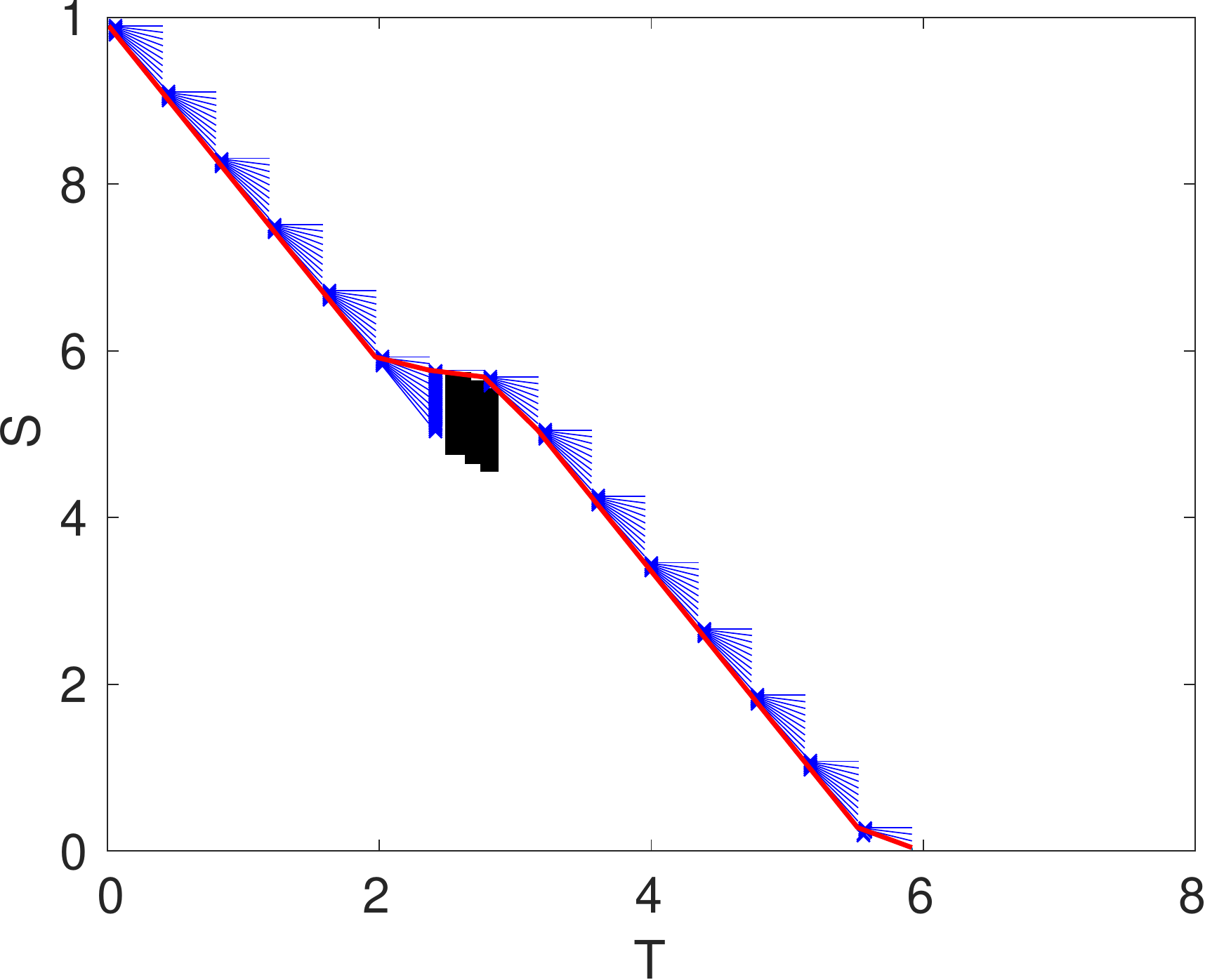}}
\subfigure[]{\includegraphics[width=.49\columnwidth]{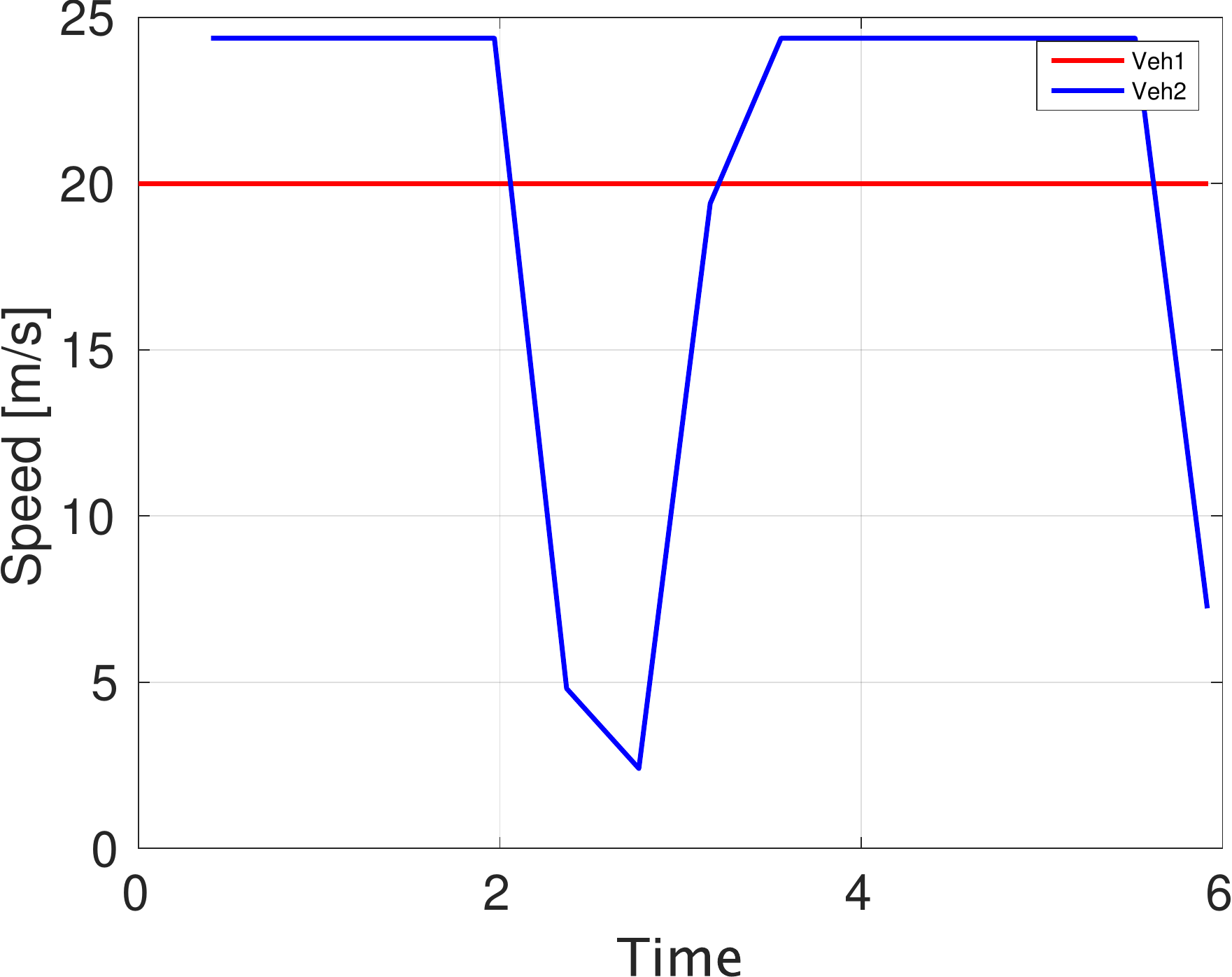}} \subfigure[]{\includegraphics[width=.49\columnwidth]{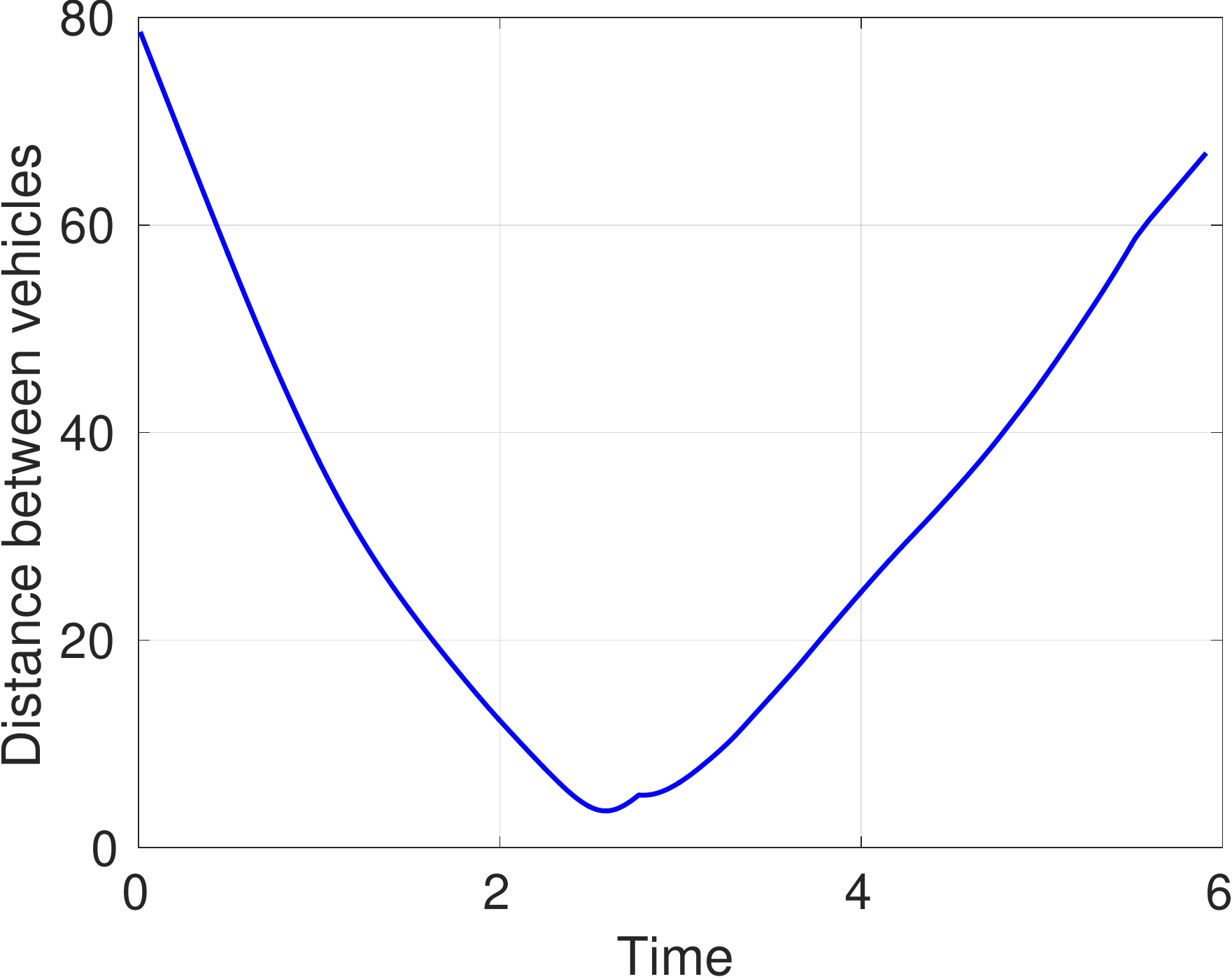}}
\caption{Simulation results of the method}
\label{Fig.Two_2}
\end{figure}
\section{Simulation Results}\label{Sec.Simulation}
\begin{figure}
\centering
\subfigure[]{\includegraphics[width=.49\columnwidth]{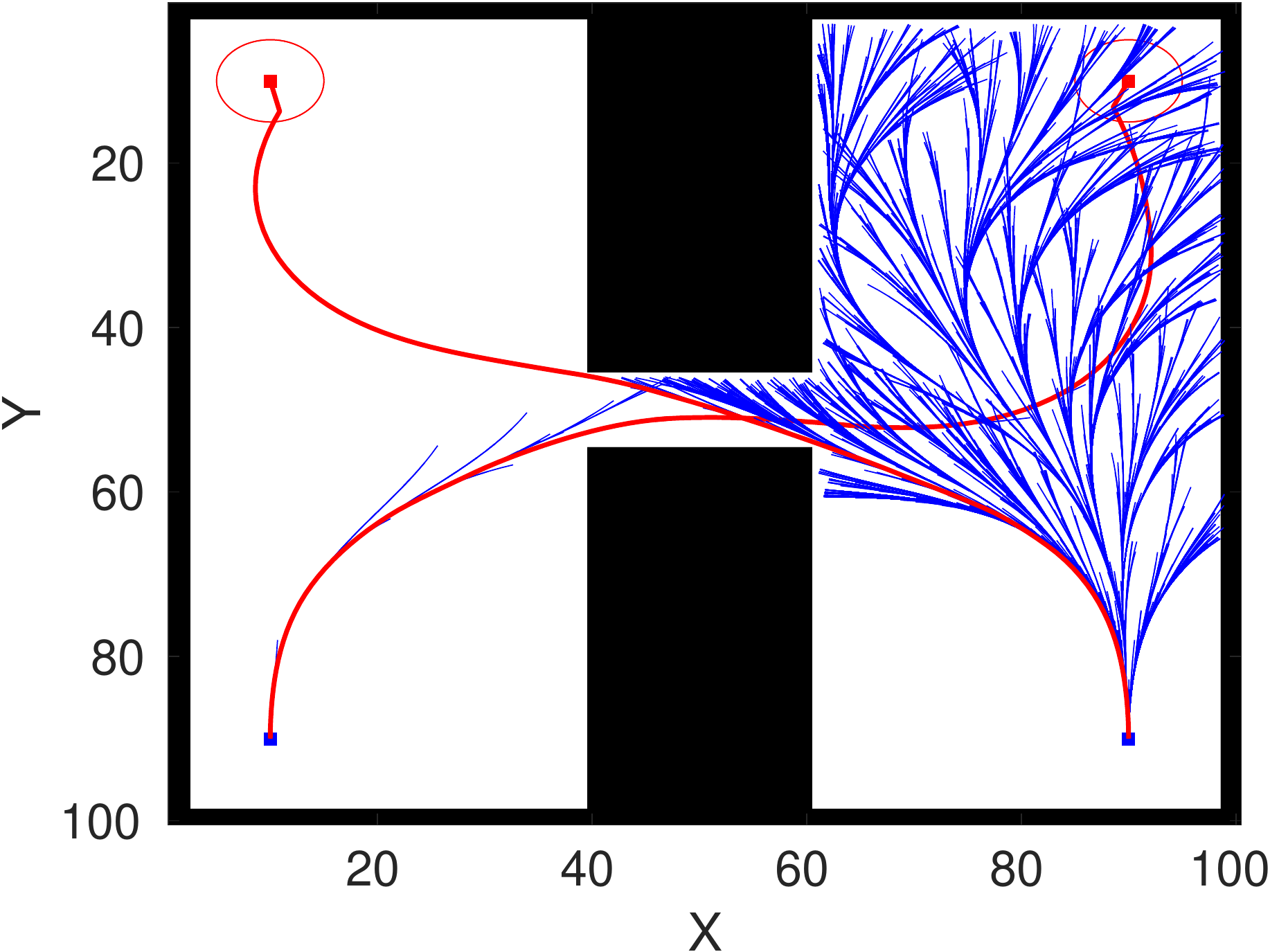}} \subfigure[]{\includegraphics[width=.49\columnwidth]{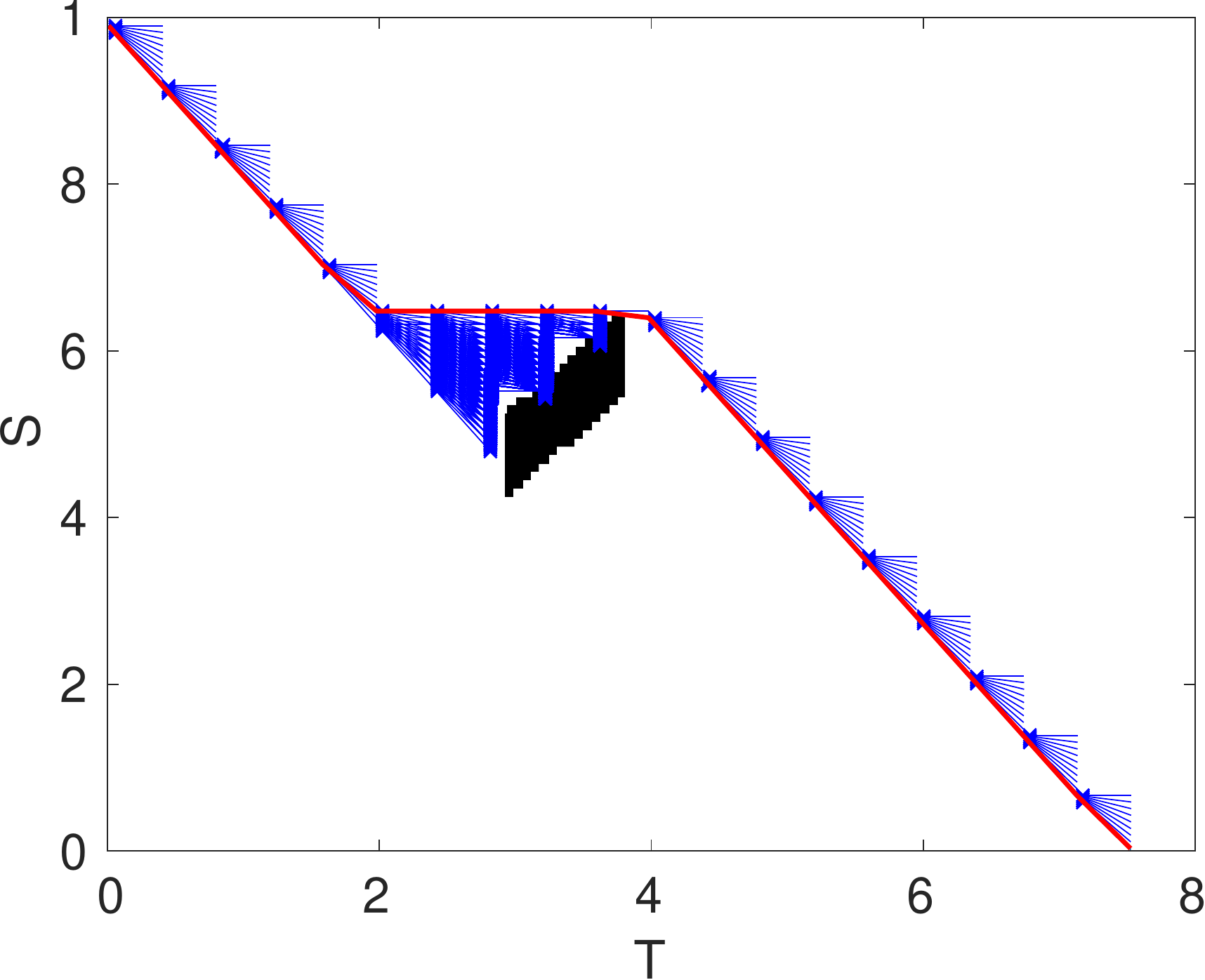}}
\subfigure[]{\includegraphics[width=.49\columnwidth]{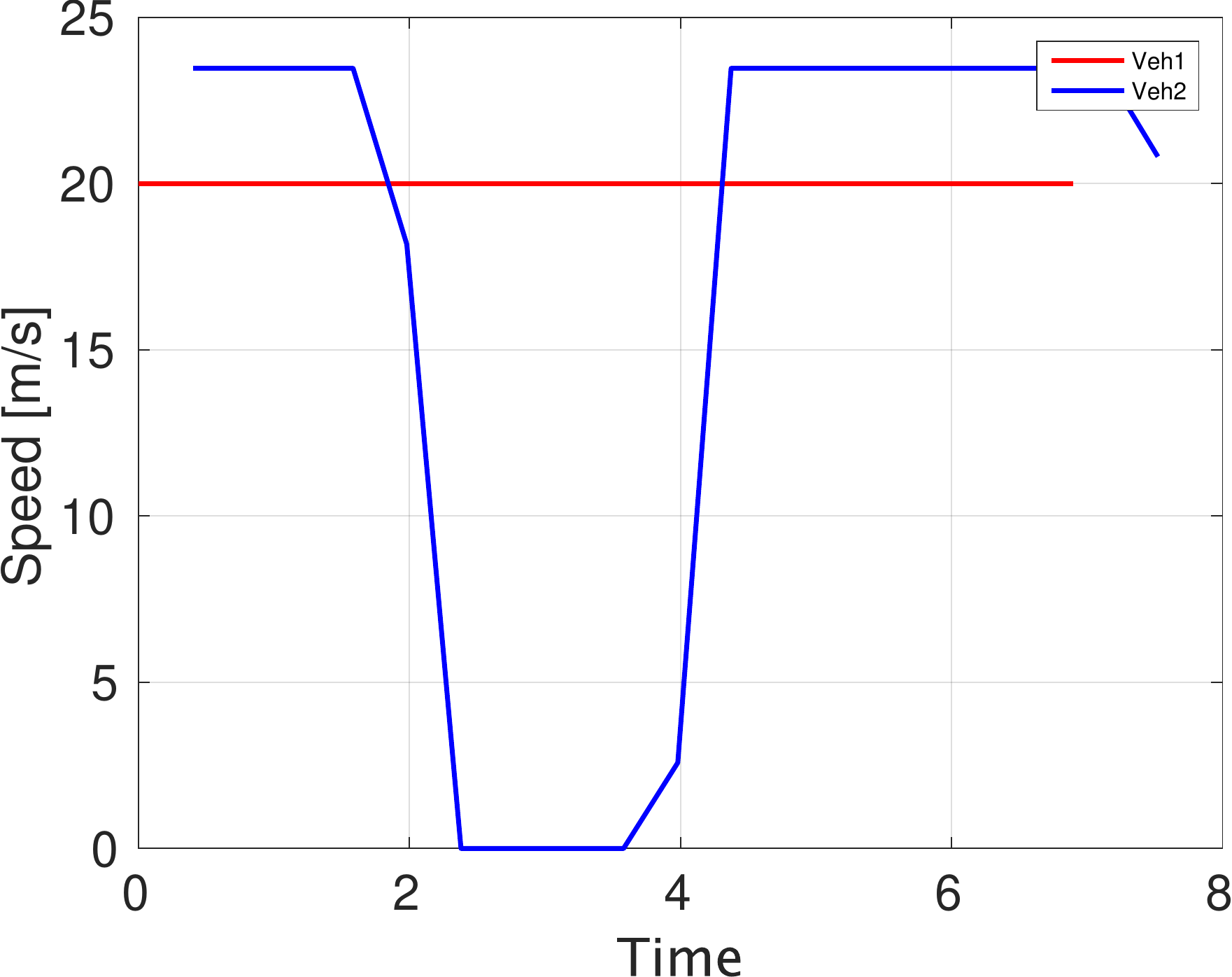}} \subfigure[]{\includegraphics[width=.49\columnwidth]{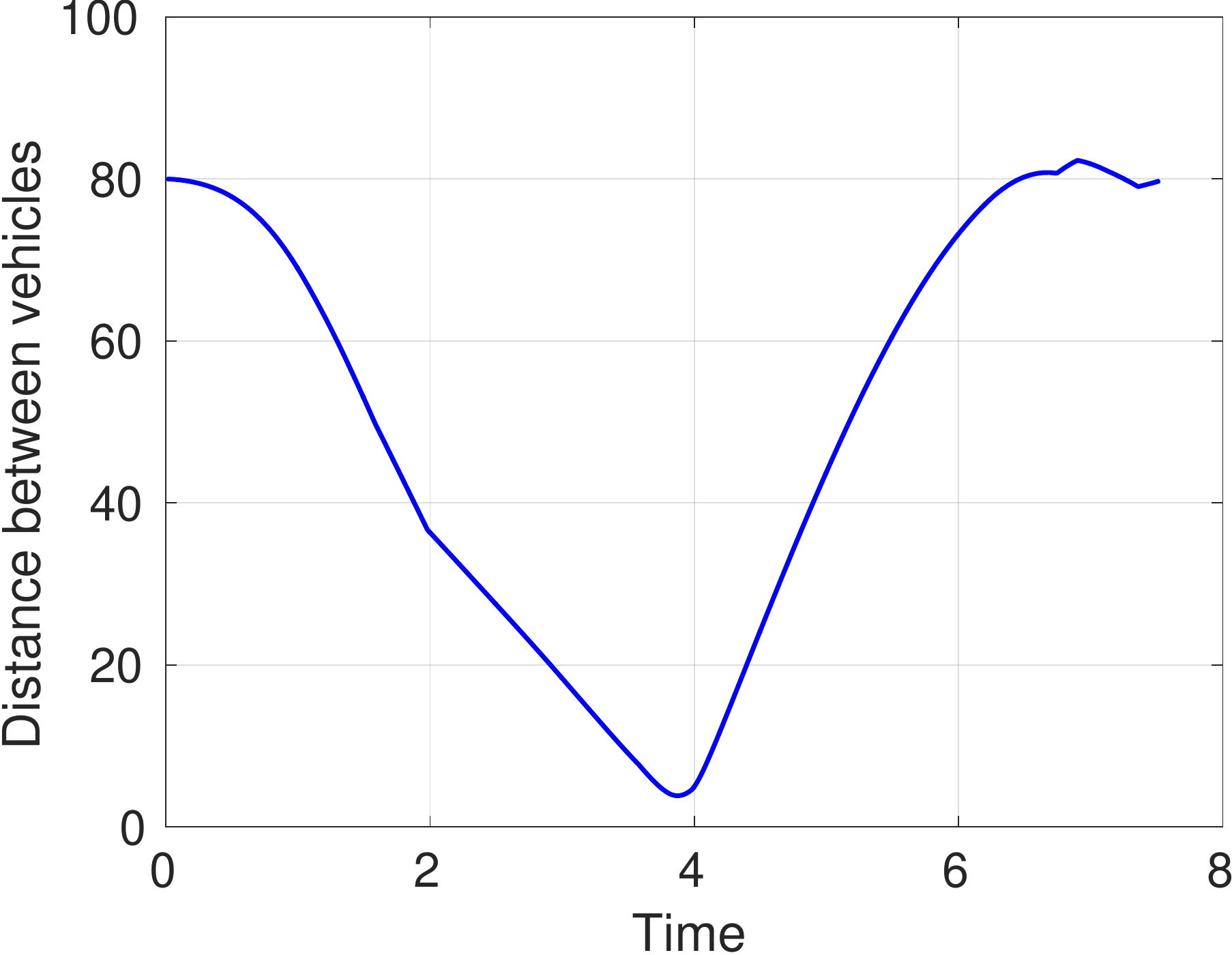}}
\caption{Simulation results of the method}
\label{Fig.Two_3}
\end{figure}

The DSBP algorithm has been tested in different scenarios. In both Fig. \ref{Fig.Two}- \ref{Fig.Two_3}, (a) illustrates the result of IRRT algorithm.
(b) is S-T map. The black blocks show the $(s,t)$ points in which there will be collision between two vehicle, blue lines show the tree branches produced by VT algorithm. Red path shows the collision free $(s,t)$ points which has minimum cost. In order to cosider vehicles diminutions, the size of the vehicles has been added to black regions. (c) shows the velocity of vehicles. The red curve illustrates the speed of highest priority vehicle. (d) shows the distance between two vehicles during the maneuver.\\
It has been observed that by using this method, vehicles can perform maneuver very fast and without collision. The maneuver times for the maneuver in obstacle rich environment was $6$ seconds and for narrow passage environment was $7$ seconds. Also, the computational time was $10$ seconds for obstacle rich environment and $11$ seconds for narrow passage environment. The results show the good performance of the algorithm in different situation including complicated obstacles that are time consuming if we want to use optimal control method.

\begin{figure}
\centering
\subfigure[]{\includegraphics[width=.49\columnwidth]{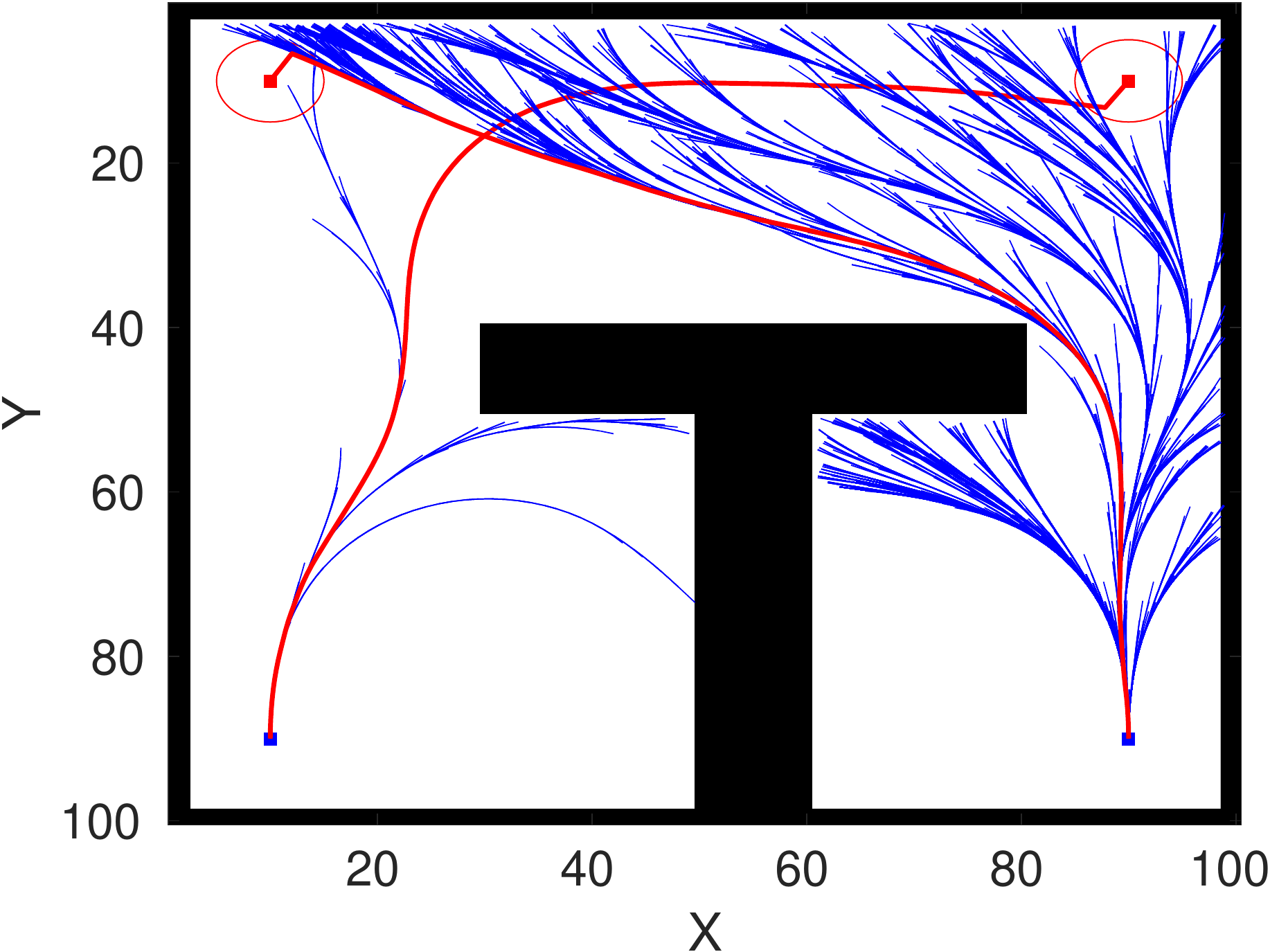}} \subfigure[]{\includegraphics[width=.49\columnwidth]{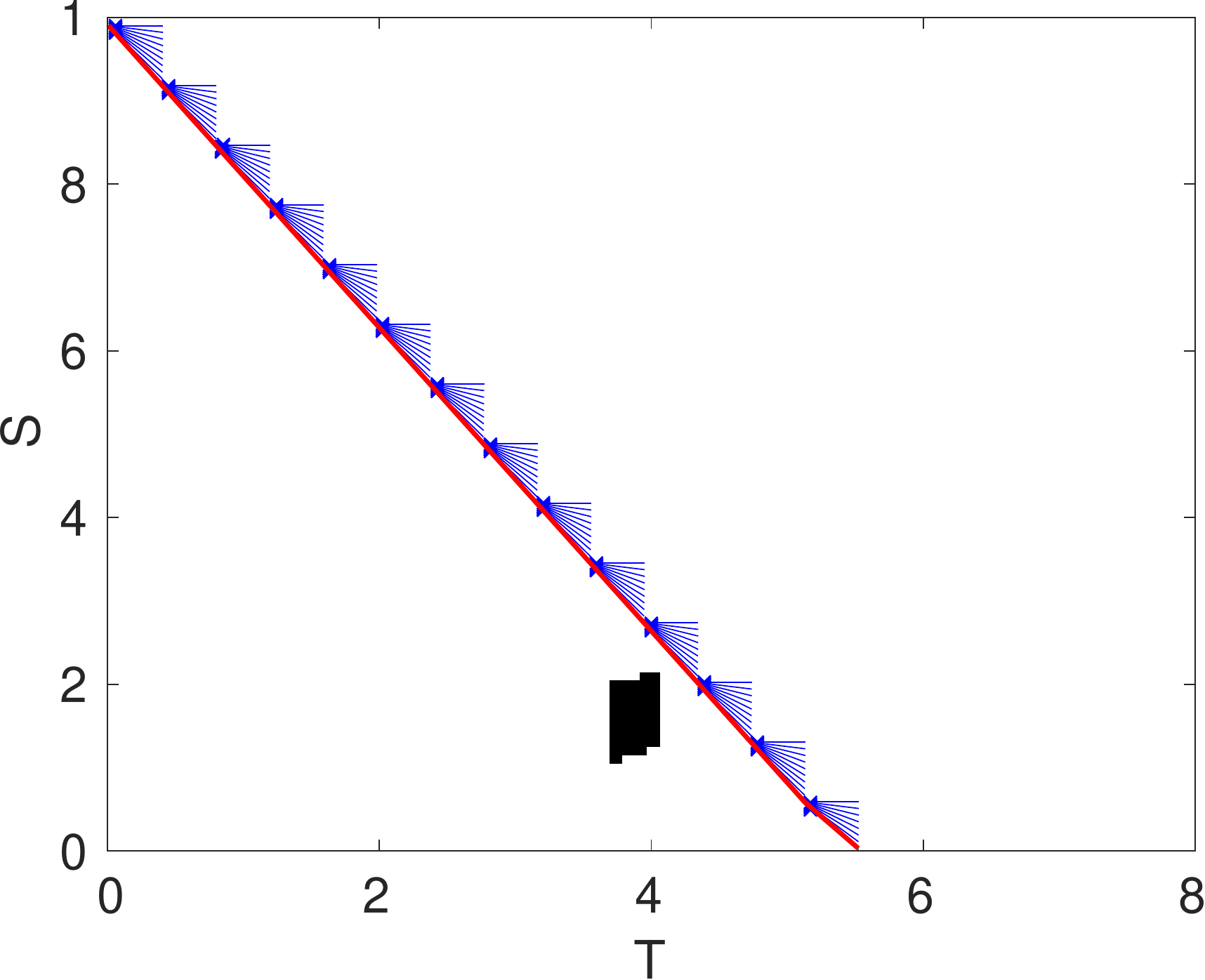}}
\subfigure[]{\includegraphics[width=.49\columnwidth]{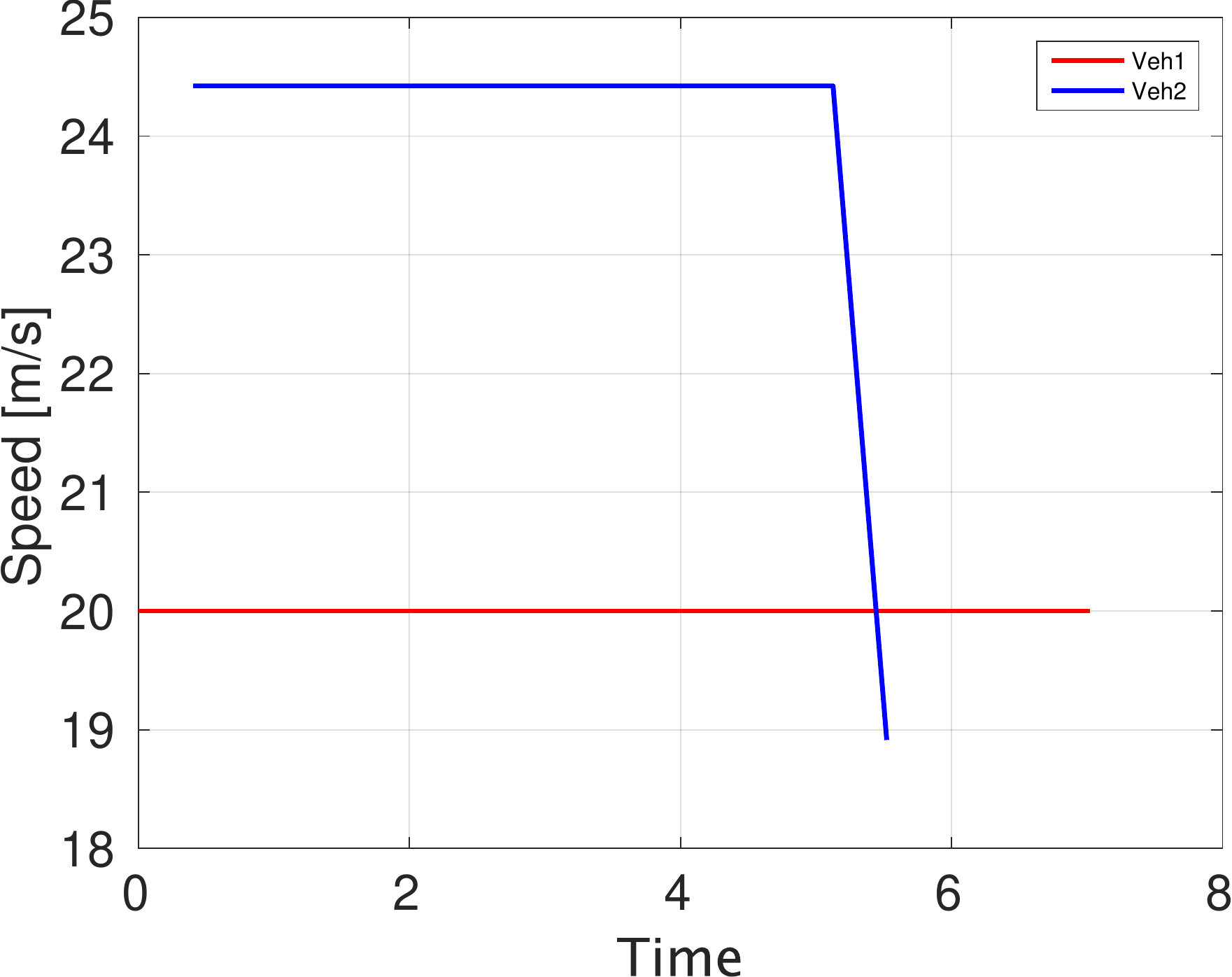}} \subfigure[]{\includegraphics[width=.49\columnwidth]{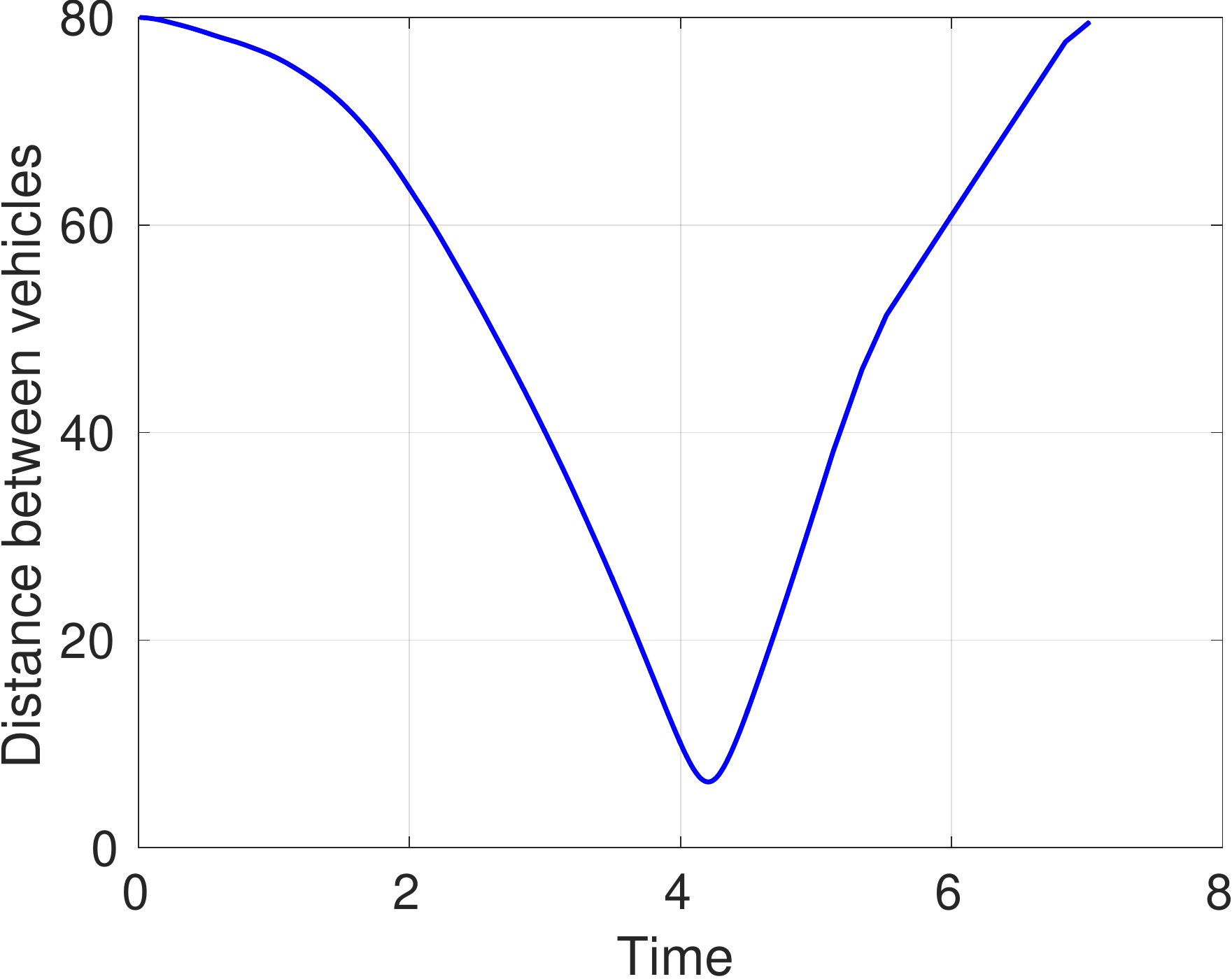}}
\caption{Simulation results of the method}
\label{Fig.Two_4}
\end{figure}

\begin{figure}
\centering
\subfigure[]{\includegraphics[width=.49\columnwidth]{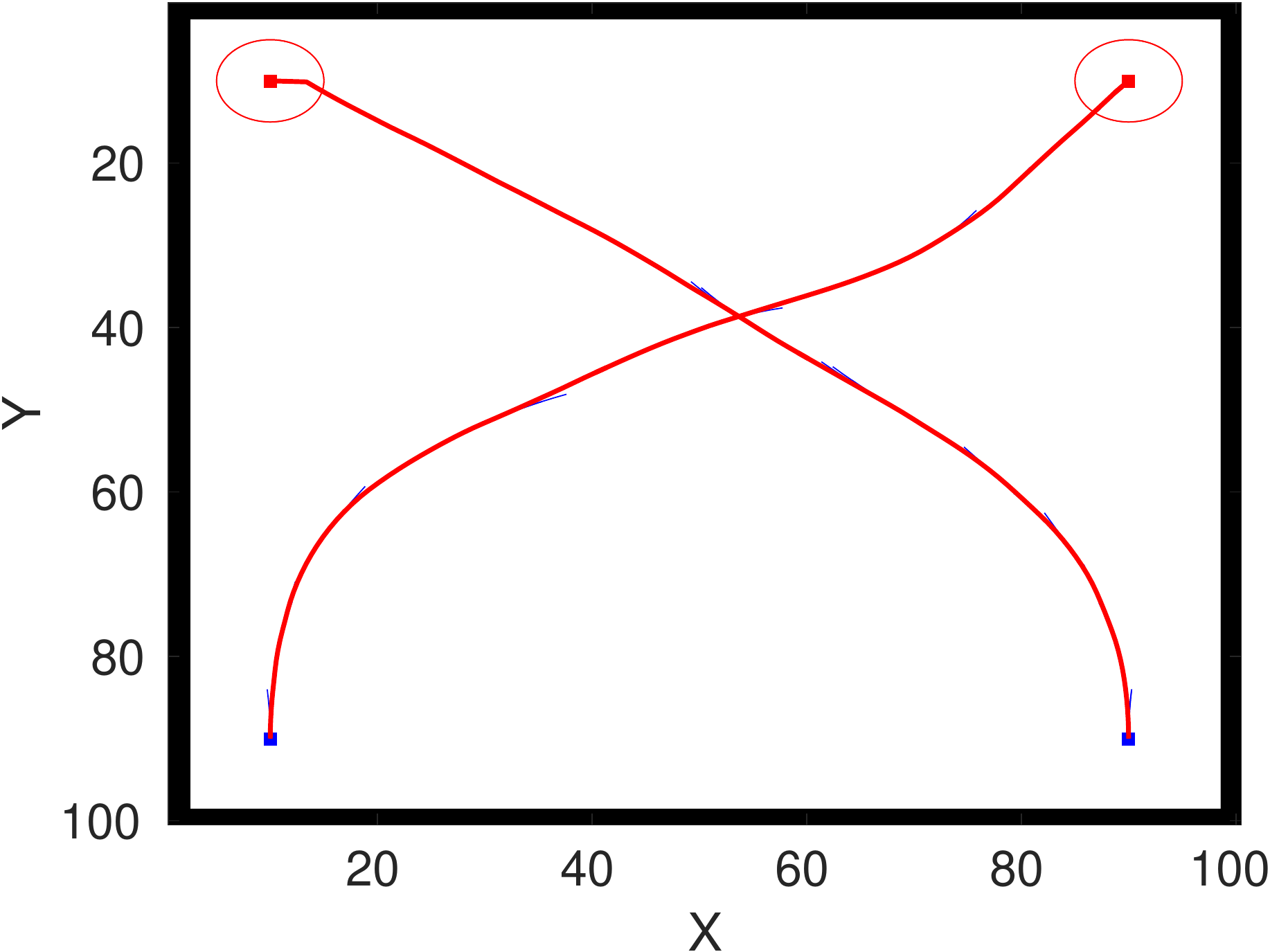}} \subfigure[]{\includegraphics[width=.49\columnwidth]{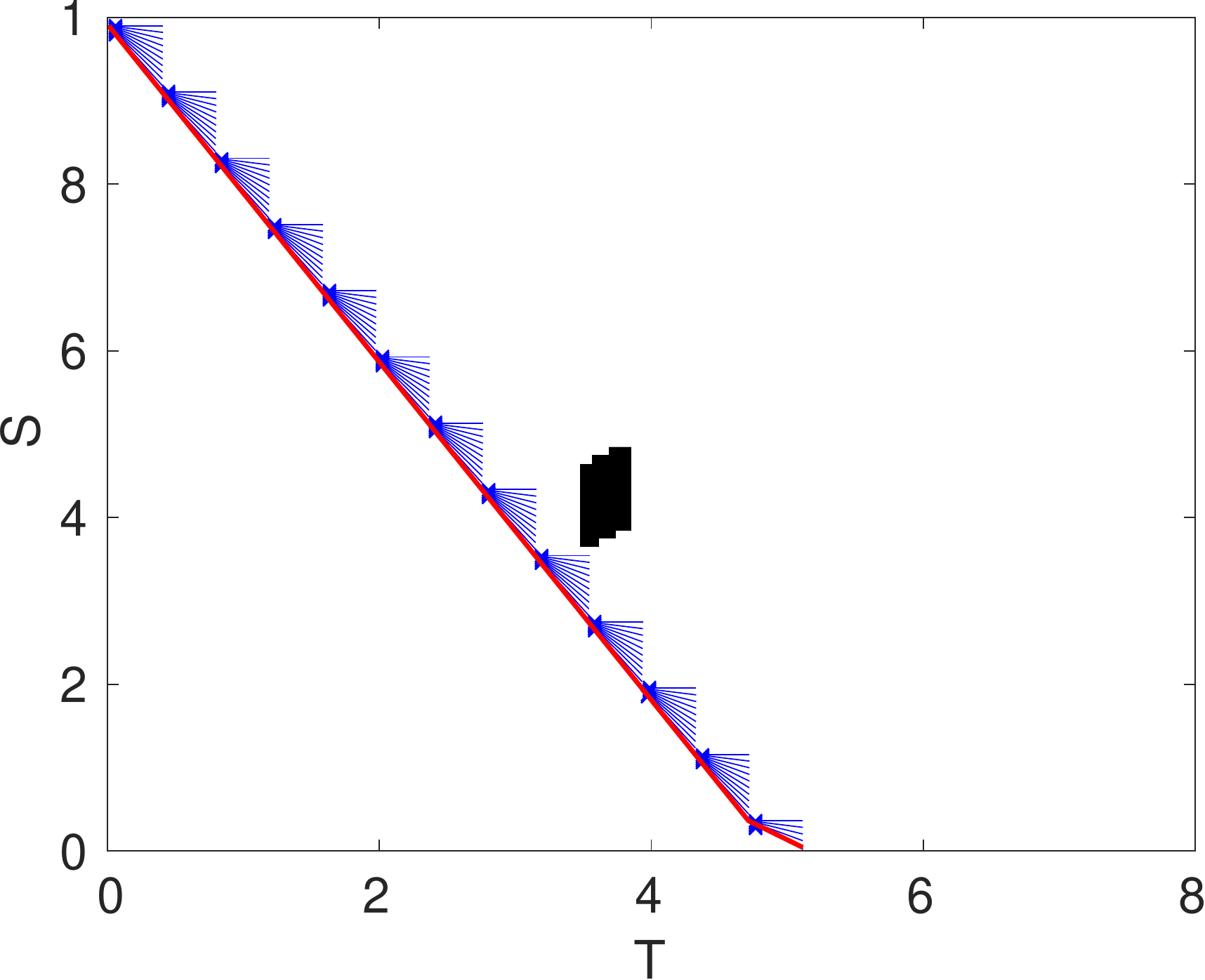}}
\subfigure[]{\includegraphics[width=.49\columnwidth]{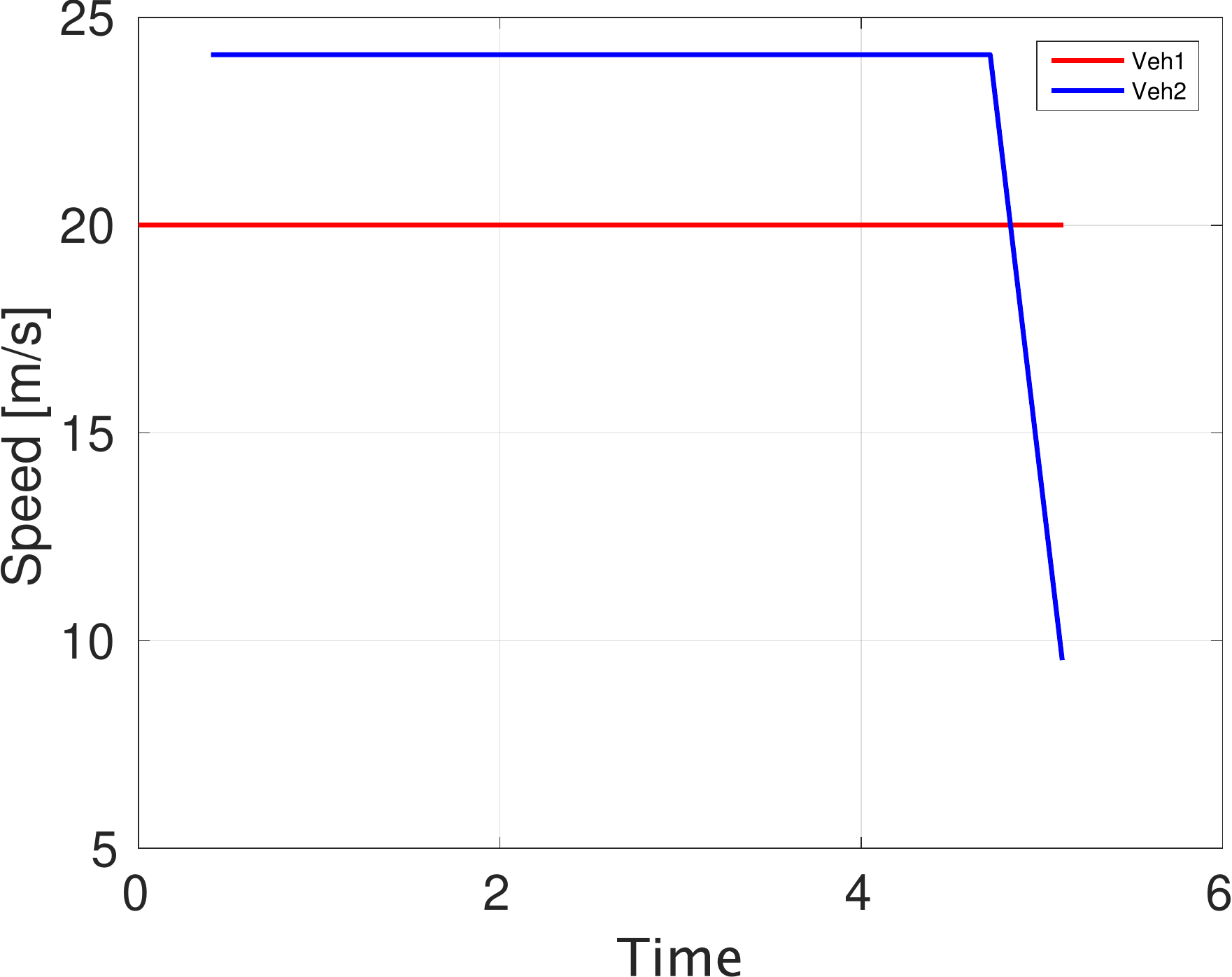}} \subfigure[]{\includegraphics[width=.49\columnwidth]{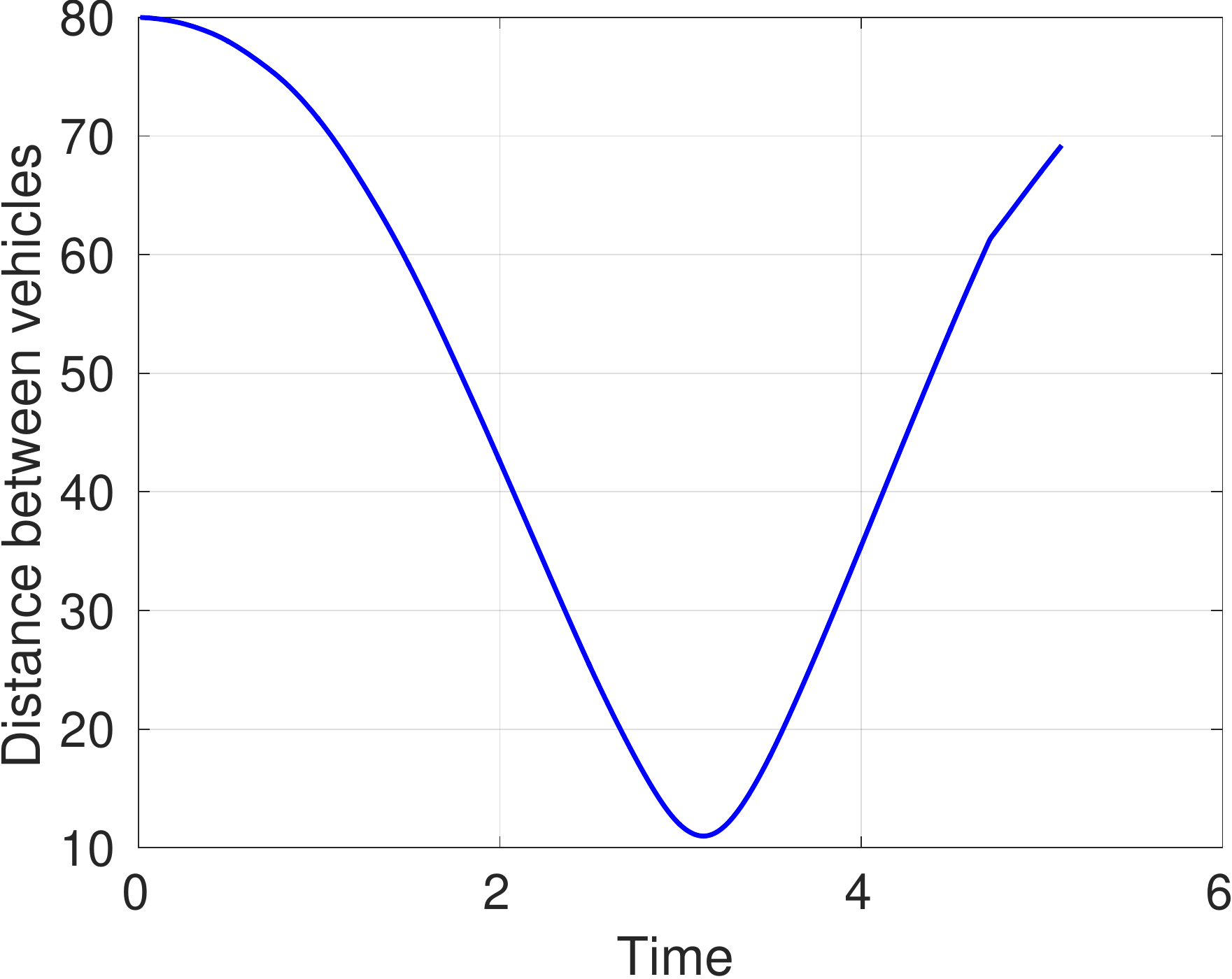}}
\caption{Simulation results of the method}
\label{Fig.Two_5}
\end{figure}

\begin{table}[hb]
  \begin{center}
  \small
  \caption{??????}\label{Tab.ODE}%
  \begin{tabular}{|c|c|c|}
  \hline
  ???& ???& ???\\\hline
  ? & ?& ?\\\hline
  ? & ?& ?\\\hline
  ? & ?& ?\\\hline
  ? & ?& ?\\\hline
  ? & ?& ?\\\hline
  ?& ?& ?\\
    ? & ?& ?\\\hline
    ? & ?& ?\\\hline
    ? & ?& ?\\\hline
    ? & ?& ?\\\hline
    ? & ?& ?\\\hline
    ?& ?& ?\\
\hline
\end{tabular}
\end{center}
\end{table}
\section{CONCLUSIONS}\label{Sec.Conclusion}
This paper proposed a sampling based planning algorithm to control autonomous vehicles. We proposed an improved Rapidly-exploring Random Tree which includes the definition of K-nearest points and proposed a two-stage sampling strategy to adjust RRT in other to perform maneuver while avoiding collision. The simulation results showed the success of the algorithm.

 {
 	\bibliography{ref}
 }

\end{document}